\documentclass[lettersize,journal]{IEEEtran}
\usepackage{amsmath,amsfonts}
\usepackage{array}
\usepackage{textcomp}
\usepackage{stfloats}
\usepackage{url}
\usepackage{verbatim}
\usepackage{graphicx}
\usepackage{cite}
\usepackage[justification=centering]{caption}
\usepackage{subcaption}
\usepackage{subcaption}   
\usepackage{wrapfig}
\usepackage[outdir=./]{epstopdf}
\usepackage{multirow}
\usepackage{makecell}
\usepackage{xcolor}
\usepackage{makecell}
\usepackage{enumitem}
\usepackage{tabularx}  

\usepackage{tikz}          
\usepackage{pgfplots}
\pgfplotsset{compat=1.18}  

\usepackage{gensymb}
\usepackage{amsmath}

\usepackage{amssymb}
\usepackage{stmaryrd}
\usepackage{amsthm} 
\usepackage[ruled,vlined]{algorithm2e}
\usepackage{algpseudocode}
\usepackage{varwidth}
\usepackage{blindtext}
\usepackage{scrextend}
\usepackage{url}
\usepackage[outdir=./]{epstopdf}

\usepackage[ruled,vlined]{algorithm2e}

\usepackage[square,comma,numbers,sort]{natbib}

\usepackage{mathtools}

\usepackage[switch]{lineno}

\usepackage{graphicx}
\usepackage{subcaption}   
\usepackage[utf8]{inputenc}   
\usepackage[T1]{fontenc}

\DeclarePairedDelimiter\floor{\lfloor}{\rfloor}

\theoremstyle{plain}

\theoremstyle{definition}

\makeatletter
\newcommand{\circlearrow}{}
\DeclareRobustCommand{\circlearrow}{%
  \mathrel{\vphantom{\rightarrow}\mathpalette\circle@arrow\relax}%
}
\newcommand{\circle@arrow}[2]{%
  \m@th
  \ooalign{%
    \hidewidth$#1\circ\mkern1mu$\hidewidth\cr
    $#1\longrightarrow$\cr}%
}
\makeatother

\usepackage{enumitem}

\begin{document}

\title{Accelerated Rotation-Invariant Convolution for UAV Image Segmentation}

\author{Manduhu Manduhu, Alexander Dow, Gerard Dooly, James Riordan 
       
\thanks{
	Manduhu Manduhu, Alexander Dow and James Riordan are with the Drone Systems Lab, School of Computing, Engineering, and Physical Sciences, University of the West of Scotland, Glasgow, Scotland. 

	Corresponding author: Manduhu Manduhu (email: manduhu.manduhu@uws.ac.uk). 
	
	Gerard Dooly is with the Centre for Robotics \& Intelligent Systems, Department of Electronic \& Computer Engineering, University of Limerick, Limerick, Ireland.
	
	}
}

\markboth{Journal of \LaTeX\ Class Files,~Vol.~xx, No.~xx, xxxx~20xx}%
{Shell \MakeLowercase{\textit{et al.}}: A Sample Article Using IEEEtran.cls for IEEE Journals}


\maketitle

\begin{abstract}

Rotation invariance is essential for precise object level segmentation in UAV aerial imagery, where targets can have arbitrary orientations and exhibit fine scale details. Conventional segmentation architectures like UNet rely on convolution operators that are not rotation invariant, leading to degraded segmentation accuracy across varying viewpoints. Rotation invariance can be achieved by expanding the filter bank across multiple orientations; however, this significantly increases computational cost and memory requirement.
In this paper, we introduce a GPU optimized rotation invariant convolution framework that eliminates the traditional data lowering (im2col) step required for matrix multiplication based convolution. By exploiting structured data sharing among symmetrically rotated filters, our method achieves multi orientation convolution with greatly reduced memory requirements and computational redundancy. We further generalize the approach to accelerate convolution with arbitrary (non symmetric) rotation angles.
Integrated into a UNet segmentation model, the framework yields up to a 5.7\% improvement in accuracy over the non rotation aware baseline. Across extensive benchmarks, the proposed convolution achieves 20–57\% faster training and 15–45\% lower energy consumption than cuDNN, while maintaining accuracy comparable to state-of-the-art rotation invariant methods. Because the scatter based operator greatly reduces intermediate feature dimensionality, the efficiency of our design also enables practical sixteen orientation convolution and pooling, yielding further accuracy gains that are infeasible for conventional rotation invariant implementations. These results demonstrate that the proposed method provides an effective and highly efficient alternative to existing rotation invariant convolution frameworks.

\end{abstract}


\section{Introduction}
\IEEEPARstart{R}{obust} 
semantic segmentation in high‑resolution aerial images requires rotation invariance, because objects of interest—whether vegetation, buildings, vehicles, or other land‑surface features—can occur at any orientation and often contain complex fine‑scale detail. Yet the convolution operator used in mainstream deep learning segmentation architectures (e.g., U‑Net) is translation equivariant but not rotation invariant, leaving performance sensitive to object orientation. Convolution is fundamental in deep learning due to its ability to efficiently extract spatial and hierarchical features from input data \cite{cnn-survey}. Convolution layers routinely account for the bulk of total FLOPs, memory traffic, and energy use in convolution‑based networks. They therefore dominate training time and bound inference throughput, making their optimization a central focus of prior works \cite{FFT1}, \cite{winograd}, \cite{10286398}.

These costs are further amplified when pursuing rotation‑invariant convolution, which typically requires evaluating kernels across multiple orientation samples.
Despite the additional computational burden, achieving rotation invariance is essential for many aerial remote sensing applications—including satellite imagery \cite{satellite2,satellite3,satellite4}, UAV imaging \cite{UAV1}, and geospatial image localization \cite{geo1}—because feature extraction must remain consistent under arbitrary transformations of the target object in the image, including rotations; this need is further underscored by recent UAV studies in geology (dual-spectrum hot-spring fluid segmentation) \cite{YI2025661}, urban scene segmentation (UAVformer) \cite{YI2023109019}, and UAV visual perception (CCTseg) \cite{YI2023112612} under varying altitudes and viewpoints.
The method in \cite{mitton2021rotation} embeds rotation equivariance into U-Net for UAV image segmentation, achieving a 7\% improvement in deforestation segmentation accuracy.

The extracted feature maps are rotation equivariant, i.e., feature maps
whose outputs are rotated in a predictable way, as we rotate the input.
A special case of equivariance is invariance \cite{harmonic-cnn}, where the output
values of the feature map are not affected by rotations of the input. 
In essence, the rotation invariance can be achieved by directly encoding rotation equivariance into the convolution layer followed by a global pooling operation. Multiple rotated versions of the canonical filter are applied at the same input to generate rotation equivariant feature maps. For instance, Cohen \& Welling \cite{G} propose a group convolution, code-name G-convolution, in which symmetric rotations (rotation angle is multiple of 90 degrees) are applied to the same filter, generating a rotation equivariant feature map. However, such rotation equivariant convolution introduces significant overhead on the computation. For example, with G-convolution, the computational overhead increases by 4 times compared to conventional convolution in 2D space, since four symmetric rotations are applied at each filter. The computational overhead possibly increases by 24 times in 3D space because there are 24 symmetric transformations available in 3D space \cite{CubeNet1}. The number of symmetric rotations and therefore the computational overhead increases along with increasing dimension of the input data. As reported in prior works \cite{11125636, yang2025group}, the increased number of symmetric rotations leads to substantial additional feature computations, making the training of rotation-invariant convolutional networks significantly slower than that of conventional convolutional models.
In this paper, we show how the proposed convolution operation can be used to accelerate rotation invariant convolution by eliminating the data lowering step and exploiting filter weight sharing.

Modern Graphics Processing Units (GPUs) play a crucial role in deep learning, providing the computational power necessary for training deep neural networks. Their parallel processing capabilities enable efficient handling of large-scale matrix operations, significantly accelerating deep learning workloads.
Matrix multiplication is an efficient method for implementing convolution on GPU, as demonstrated by \cite{cuDNN}. 
This requires transforming the input image into matrices that are suitable for fast multiplication.
This transformation can be achieved by lowering the input data (im2col) with duplication.
The GPU implementation of the proposed method in this paper is also based on fast GPU matrix multiplication, but without data lowering. 

As described in the previous work \cite{manduhu2025airbornesensedetectdrones}, conventional (gather‑style) convolution accumulates contributions at each output location: every neighboring input pixel is multiplied by its corresponding kernel weight, and the results are summed into the central output position. In contrast, our scatter convolution reverses this data flow: each input pixel is multiplied by a bank of filter weights and the resulting products are scattered directly to their destination output locations. We originally introduced this formulation to support sparse convolutions on irregularly sampled LiDAR data. In this paper, we generalize the scatter approach to (i) dense, standard grid convolution and (ii) rotation‑invariant convolution. The scatter mapping removes the im2col‑style data‑lowering step typically required by matrix multiplication based implementations, reducing memory traffic and simplifying the kernel. Moreover, because the scatter mapping exposes weight sharing across symmetric rotations, intermediate results can be reused—collapsing the number of distinct multiplications to one per symmetry group and substantially lowering computational cost.

While computational efficiency is an important consideration, achieving rotation invariance in UAV image segmentation inherently involves a trade-off between segmentation accuracy and computational efficiency. Increasing the number of modeled orientations generally improves accuracy by enhancing rotational robustness and feature expressiveness; however, this improvement comes at the cost of increased computation, memory traffic, and intermediate feature-map growth. As demonstrated in our experiments, moderate orientation resolutions (e.g., eight orientations) provide a favorable balance by significantly improving segmentation accuracy while remaining computationally feasible. In contrast, further increasing the orientation resolution leads to prohibitively large intermediate feature maps for conventional rotation-invariant implementations, making them infeasible to execute within GPU memory constraints. This work therefore focuses not only on improving computational efficiency, but also on enabling practical high-resolution rotation handling by reducing redundant computation and memory overhead, allowing finer orientation sampling under realistic GPU constraints.

\textcolor{black}{Our main contributions are as follows:}
\begin{itemize}
\item \textcolor{black}{\textbf{GPU-optimized single-orientation convolution:} Since a single-orientation convolution reduces to the standard convolution, we propose an efficient GPU implementation of standard convolution using matrix multiplication without data lowering (im2col), which significantly reduces computational operations and memory access overhead.}

\item \textcolor{black}{\textbf{Symmetric rotation optimization:} For symmetric rotations (90°, 180°, 270°, and 360°), we exploit data sharing via scatter operations, allowing each multiplication to be computed once and reused across all symmetric orientations.}

\item \textcolor{black}{\textbf{Arbitrary rotation acceleration:} The framework is further generalized to support arbitrary rotation angles, achieving efficient convolution under continuous rotational transformations.}

\item \textcolor{black}{\textbf{Comprehensive evaluation:} Integrated into a UNet model, the proposed convolution yields up to a 5.7\% improvement over the non rotation aware baseline, while achieving 20–57\% faster training and 15–45\% lower energy consumption than cuDNN across all evaluated workloads. Its efficient scatter based design also enables practical sixteen orientation convolution, providing additional gains unattainable with conventional rotation invariant methods.}
\end{itemize}

\section{Related Work} \label{ref}

\subsection{Segmentation of remote sensing imagery} \label{subsec:seg-rsd}
Early semantic segmentation models were largely based on convolutional neural networks (CNNs) \cite{7789580}. A key milestone was the fully convolutional network (FCN) introduced in \cite{FCN}, which adopts an encoder–decoder architecture composed entirely of convolutional layers, without any fully connected layers. Building on this foundation, \cite{rs13234902} proposed an improved FCN framework for remote sensing image segmentation, incorporating multi-scale feature extraction, enhanced encoder–decoder connections, and refined upsampling to better handle complex, high-resolution imagery.
To obtain larger receptive fields without lowering image resolution via downsampling, DeepLab was introduced in \cite{deeplabv1}, utilizing atrous convolutions to achieve dense, multi-scale feature extraction. The work in \cite{10608051} further improves DeepLab by aggregating multi-scale features to enhance aerial image segmentation. Many early segmentation models did not fully exploit low-level features in the decoder, resulting in the loss of fine spatial detail. U-Net \cite{ronneberger2015u} addresses this limitation by introducing skip connections that link encoder and decoder features, enabling more effective fusion of detailed and high-level information. U-Net and its variants have also been widely adopted for UAV and aerial image segmentation, as in UVid-Net for UAV video segmentation \cite{9392319} and Res-U-Net-based coastal boundary detection from high-resolution UAV imagery \cite{wang2025multi}.

Attention-based architectures have also been used for aerial image segmentation. SegHSI \cite{10751785} adopts a CNN-based backbone augmented with cluster attention to model spectral–spatial correlations in hyperspectral imagery. The work in \cite{10422983} employs a Transformer–CNN hybrid encoder–decoder for UAV forestry segmentation, enabling accurate delineation of diseased pine canopies in high resolution imagery.

\subsection{Rotation invariance in remote sensing imagery} \label{subsec:seg-rsd}
Rotation invariance (or equivariance) is crucial for accurate object-level segmentation in UAV imagery, as targets often appear in arbitrary orientations.
The work in \cite{isprs-annals-V-1-2022-15-2022} introduces a CNN-based framework for vehicle instance segmentation in UAV images that uses rotated bounding boxes instead of standard axis-aligned boxes to better match the true orientation of vehicles.
The method in \cite{9466361} learns rotation-invariant deep embeddings by aligning features from multiple rotated views of the same image, enabling robust analysis of remote sensing images under arbitrary orientations. FRINet, introduced in \cite{10339376}, is built on CNN feature extractors and incorporates a rotation-adaptive matching mechanism to enable few-shot, rotation-invariant segmentation of aerial images. The paper \cite{mo2024achieving} proposes a mechanism-assured convolution that embeds rotation invariance directly into the operation itself, avoiding reliance on data augmentation and ensuring consistent feature responses under rotated inputs.
RIC-CNN \cite{mo2024ric} achieves rotation invariance by embedding polar coordinate information into convolution operations, enabling the network to generate stable feature representations under arbitrary rotations.
The method in \cite{mitton2021rotation} embeds rotation equivariance into UNet by using the discrete rotation group $C_8$ ($45^\circ$ increments) as the symmetry group, enabling stable feature representations that improve both deforestation segmentation and driver classification under arbitrary image orientations.
In this paper, we introduce a computationally efficient rotation-invariant convolution and incorporate it into a U-Net architecture for semantic segmentation of UAV imagery. 

\subsection{Single-orientation rotation convolution}\label{sec:dense-conv} \label{subsec:standard-conv}
As single-orientation convolution reduces to standard convolution, we review efficient implementations of standard convolution in this section. Several FFT-based approaches \cite{FFT1,FFT2,FFT3,FFT4} have been proposed to reduce the computational cost for standard convolution on different hardware. FFT-based approaches can significantly lower the work complexity of convolution, but it is only effective for convolution with larger filters \cite{cuDNN}. 

Lavin \& Gray \cite{Lavin} proposed a method based on the Winograd algorithm \cite{winograd} for convolution with small filters. The key idea of Winograd-based convolution is similar to the one based on FFT which performs multiplications in the complex domain. However, unlike FFT-based convolution, the Winograd-based convolution operates on real numbers, thus requiring fewer operations. 
For a filter with larger size, Winograd-based convolution shows poor performance, since it needs to perform a significant amount of extra arithmetic such as additions and data transformations.

To reduce the substantial memory usage in convolution with matrix multiplication, Lu \textit{et al.} \cite{lu2023im2win} proposed new data-lowering techniques. However, their method still requires data duplication. In \cite{10.5555/3213069.3213072}, Chang \textit{et al.} proposed an approach to improve memory efficiency in convolution operations by effectively utilizing the on-chip memory of GPUs. In \cite{9235051}, a matrix multiplication-based approach without data lowering is proposed. However, it requires recomputing writing positions, preventing direct pointwise operations.

\subsection{Rotation-equivariant convolution for multi-orientation rotations} \label{subsec:rot-conv}
A general theory of an equivariant CNN can be found in \cite{cohen2018general}. The mathematical formulation of equivariance through a weight sharing scheme is exhibited in \cite{equi-theory}.

G-convolution \cite{G} applies a group of filters to each feature map, where each filter is a transformed version of the original filter. Two different groups are proposed in \cite{G}, code-name $p4$ group and $p4m$ group. The $p4$ group consists of 4 filters where each filter is obtained by rotating the original filter by 0 degrees, 90 degrees, 180 degrees and 270 degrees, respectively. The $p4m$ group consists of 8 filters which are obtained by combining $p4$ group and reflection transformations. The convolution with $p4$ group or $p4m$ group will increase the computational overhead by 4 or 8 times. 
There are multiple works \cite{CubeNet1, CubeNet2} applying the G-convolution on 3D volume data. In \cite{CubeNet1}, different groups code-name \textit{Cube Group}, \textit{Klein's Four-group} and \textit{Tetrahedral Group} are proposed. Among these groups, the largest group is the \textit{Cube Group} which consists of 24 filters resulting in an increase of computational overhead by 24 times. Obviously, the number of filters in a group will increase along with increasing dimension of the input data and therefore the computational overhead. 
Hexagonal group convolution \cite{hexa} rotates a filter by any multiple of 60 degrees, without interpolation. They define the following two hexagonal group convolutions: groups $p6$ and $p6m$, which contain integer translations (orientation cycling), rotations with multiples of 60 degrees, and mirroring for $p6m$. The paper \cite{G-cnn10} proposes Gauge-equivariant convolutional network for manifolds, which uses Hexagonal group convolution to implement the weight sharing scheme for Gauge-equivariant. 
The work in \cite{shallowNet}
proposes a shallow network which
performs convolution with rotated versions
of each canonical filter, where the rotation is with arbitrary angle.
The convolution is followed by a global pooling over orientations.
The number of rotations applied is 32 i.e. the angle increment between 
successive two rotations is $\frac{\pi}{16}$.
Bicubic interpolation is used for sampling after rotation.

Oriented Response Network (ORN) \cite{orn} introduces active rotating filters that generate orientation-aware feature maps, allowing CNNs to model rotations explicitly and achieve rotation invariance or equivariance without heavy data augmentation.
The work in \cite{e2cnn} provides a unified framework for designing steerable CNNs that are equivariant to all
$E(2)$ transformations and offers the widely used E2CNN library for generating rotation-equivariant features.

Filters with steerability can be
reconstructed at any orientation as a finite, linear combination of base
filters. This eliminates the need to learn multiple filters at different
orientations.
Harmonic Network proposed in \cite{harmonic-cnn} 
achieves rotation equivariance in continuous rotations by  
learning coefficients of base filters, where
these base filters are used to form the filters (steerable) of the CNN model.
In the network proposed in \cite{G-cnn7},
rotation equivariance is guaranteed by using 
the G-convolution in which the number of rotation is more than four.
The rotated filter (steerable) in a group is constructed by manipulating coefficients of the base filter. 
In this way, it avoids the use of interpolation which usually introduces artifacts in the rotated filter.  
The network shown in \cite{gabor} achieves rotation equivariance by multiplying each convolutional filter by 
several oriented Gabor filters \cite{weldon1996efficient}. In the back propagation stage, only the convolutional filters are updated.
The method in \cite{cheng2018rotdcf} reduces the number of parameters and computational complexity of rotation-equivariant
CNNs by decomposing convolutional filters (steerable) under joint
steerable bases over space and rotations simultaneously. Meanwhile, the performance is preserved. 

Rotation equivariance typically leads to a rapid increase in feature dimensionality as the number of sampled orientations grows. To tackle this problem of exploding dimensionality, a rotation equivariant convolution is usually followed by an orientation pooling operation, after which the layer becomes rotation invariant. The orientation pooling can be implemented in two different ways: point-wise average pooling \cite{featureTrans} or point-wise max pooling \cite{shallowNet}.

In this paper, we describe how our proposed convolution operation is applied to accelerate a rotation invariant convolution layer, which combines symmetric rotation and steerable transformations. 

\section{Methodology} \label{sec:methodology}

\textcolor{black}{
UAV aerial image segmentation presents unique challenges compared with ground-level imagery, as target objects often appear at arbitrary orientations due to the aerial imaging perspective. Conventional convolution operators are sensitive to such orientation variations, which can lead to degraded segmentation accuracy. The methodology presented in this section introduces a series of convolution formulations that are specifically designed as drop-in replacements for standard convolution layers in semantic segmentation networks.These operators form the computational backbone of the proposed UAV segmentation framework, enabling rotation-aware feature extraction while maintaining efficiency suitable for large-scale aerial imagery. Their practical effectiveness is demonstrated through integration into a U-Net architecture and validated in the experimental section.
}

\subsection{Single orientation convolution formulation}
\begin{figure}[h]
\centering
    \includegraphics[width=8.5cm]{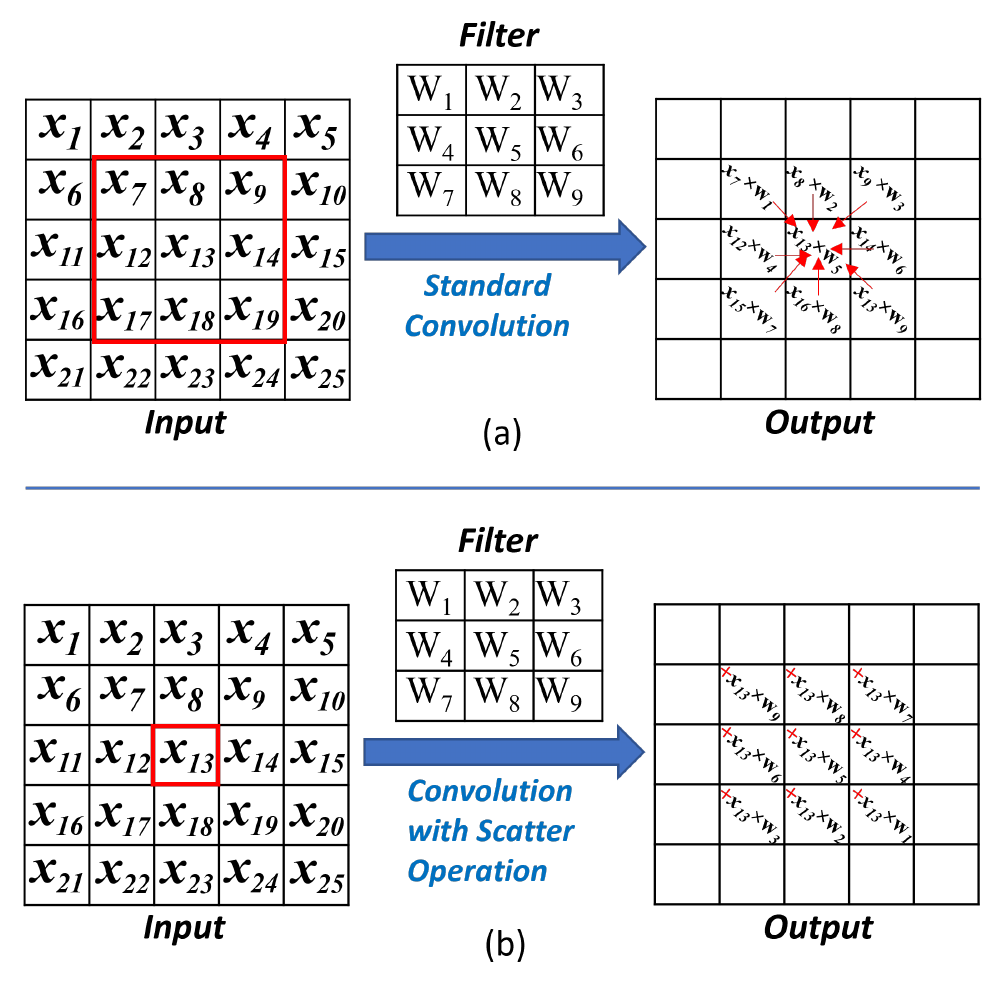}
\caption{(a) Standard convolution where $x_i, w_i, y_i$ represent input, filter and output, respectively. \textcolor{black}{(b) Convolution with scatter operation where each input element is multiplied by each filter element and the results are added with those of neighboring elements.} }
\label{fig:standardConv}
\end{figure}

In the single-orientation case, the operation reduces to standard convolution because only one fixed kernel orientation is applied and no orientation-dependent processing is required. In standard convolution, each filter element is multiplied by its overlapping input element, and the resulting products are summed to produce the output value at the corresponding spatial location, as illustrated in Fig.~\ref{fig:standardConv}(a). Alternatively, as demonstrated in our previous work~\cite{manduhu2025airbornesensedetectdrones}, convolution can also be performed using a scatter-based approach (which we employed to implement sparse convolution). In this approach, each input pixel is multiplied by different kernel elements, and the resulting values are added to their corresponding neighboring output positions, as shown in Fig.~\ref{fig:standardConv}(b). This distributes computation across output locations rather than accumulating it at a single pixel, enabling convolution without data lowering while benefiting from the high-throughput matrix multiplication capabilities of modern GPUs. In this paper, we extend the scatter-based approach to support dense convolution.

Let the input be 
$\mathbf{X} \in \mathbb{R}^{C_{\mathrm{in}} \times H \times W}$ 
and the filters 
$\mathbf{W} \in \mathbb{R}^{C_{\mathrm{out}} \times C_{\mathrm{in}} \times K_h \times K_w}$, where 
$C_{\mathrm{in}}$ and $C_{\mathrm{out}}$ denote the numbers of input channels and output
channels, respectively, and 
$H$ and $W$ represent the height and width of the input feature map,
while $K_h$ and $K_w$ denote the kernel height and width.
For valid convolution, the output spatial size is
\[
H' = H - K_h + 1,
\qquad
W' = W - K_w + 1.
\]
The convolution output is
\begin{equation}
\begin{split}
Y_{c_{o},h,w}
=
\sum_{c_i=0}^{C_{\mathrm{in}}-1}
\sum_{i=0}^{K_h-1}
\sum_{j=0}^{K_w-1}
W_{c_o,c_i,i,j}\;
X_{c_i,\,h+i,\,w+j}, \\
h  \in \{0,...,H'-1\}\\
w  \in \{0,...,W'-1\}\\
c_o  \in \{0,...,C_{out}-1\}
\end{split}
\end{equation}
which is the standard sliding-window inner product formulation.
To convert convolution into matrix multiplication, every 
$K_h \times K_w$ receptive field from $\mathbf{X}$ is flattened into a column, forming
$\mathbf{X}_{\mathrm{col}} \in \mathbb{R}^{(C_{\mathrm{in}} K_h K_w)\, \times\, (H' W')}.$
Formally,
\begin{equation}
X_{\mathrm{col}}(k,t)
=
X(c_i,\, h+i,\, w+j),
\end{equation}
where
\[
k = c_i K_h K_w + i K_w + j,
\qquad
t = h W' + w,
\]
and $h$ and $w$ enumerate output spatial locations 
($h \in \{0,\ldots,H'-1\},\; w \in \{0,\ldots,W'-1\}$), 
while $i$ and $j$ enumerate positions inside the 
$K_h \times K_w$ receptive field 
($i \in \{0,\ldots,K_h-1\},\; j \in \{0,\ldots,K_w-1\}$). 
Thus $X(c_i,\,h+i,\,w+j)$ selects the input element covered by the kernel
at offset $(i,j)$ for input channel $c_i$.
Each \emph{column} corresponds to one sliding patch, and each \emph{row} indexes one element in the flattened receptive field.

Each filter kernel is flattened into a row vector:
\[
\mathbf{W}_{\mathrm{row}} 
\in 
\mathbb{R}^{C_{\mathrm{out}} \times (C_{\mathrm{in}} K_h K_w)},
\]
with
\[
W_{\mathrm{row}}(c_o,k)
=
W(c_o,c_i,i,j).
\]
The convolution can now be computed as a single matrix multiplication:
\[
\mathbf{Y}_{\mathrm{col}}
=
\mathbf{W}_{\mathrm{row}}\,
\mathbf{X}_{\mathrm{col}}.
\]
Clearly, flattening the receptive fields of $\mathbf{X}$ causes each input element to be replicated \(K_h K_w\) times, i.e.,
$\mathbf{X}_{\mathrm{col}}$ contains up to $K_h K_w$ copies of each element of $\mathbf{X}$. 
Consequently, the number of multiplications increases by a factor of \(K_h K_w\).

For a single-channel input and a single filter, 
the general formula of the convolution with scattering operation is given as follows: 
\begin{equation}
\begin{split}
Y_{h - m +  \floor {K_h/2}, w - n + \floor {K_w /2}} \; += X_{h,w} \times W_{m, n} \\
m \in \{0, ..., K_h-1\}, n \in \{0, ..., K_w-1\} \\
h  \in \{0,  ...,H'-1\}, w  \in \{0, ...,W'-1\}\\
\end{split}
\end{equation}

For $C_{\mathrm{in}}$ input channels and $C_{\mathrm{out}}$ output channels, the convolution with scattering can be expressed as follows: 
\begin{equation}
\label{eq:multi_ch_scatter}
\begin{split}
Y_{c_{o}, i - m +  \floor {K_h/2}, j - n + \floor {K_w /2}} \; += \\
\sum_{c_i=0} ^{C_{in}-1}{X_{c_{i}, h,w} \times W_{c_{o}, c_{i}, m, n}} \\
m \in \{0, ..., K_h-1\}, n \in \{0, ..., K_w-1\} \\
h  \in \{0,  ...,H'-1\}, w  \in \{0, ...,W'-1\}\\
c_o  \in \{0,...,C_{out}-1\}
\end{split}
\end{equation}
As shown on the right-hand side of Formula~\eqref{eq:multi_ch_scatter}, the matrix multiplication is used to perform channel-wise multiplication and summation. 
The results are then added to the corresponding output positions following the scatter rules.
This approach eliminates the need for data duplication when performing convolution using matrix multiplication.

The pseudocode for single-channel scatter convolution is shown in Algorithm~\ref{algorithm1}.
The number of multiplications and additions required is the same as standard convolution, it is $O(H \times W \times K_h \times K_w)$.
\begin{algorithm}
\SetAlgoLined
\KwIn{An image $X(i,j)$ of size $H \times W$ and a filter $W(m,n)$ of size $K_h \times K_w$}
\KwOut{A new image $Y(x,y)$ after convolution}
\For{$i \leftarrow 0$ \KwTo $H-1$}{
  \For{$j \leftarrow 0$ \KwTo $W-1$}{
    \For{$m \leftarrow 0$ \KwTo $K_h-1$}{
      \For{$n \leftarrow 0$ \KwTo $K-w-1$}{
	$x \leftarrow i - m +  \floor {K_h/2}$\;
	$y \leftarrow  j - n + \floor {K_w /2}$\;
        $Y(x,y) \leftarrow Y(x,y) + X(i, j) \times W(m,n)$\;
      }
    }
  }
}
\Return{$Y(x,y)$}\;
\caption{Convolution with scatter operation}
\label{algorithm1}
\end{algorithm}

\subsection{Symmetric rotation-invariant convolution formulation} \label{subsec:accSymRot}

Symmetric rotation equivariant can be achieved by rotating a filter under the discrete rotational symmetries of $0^\circ$, $90^\circ$, $180^\circ$, and $270^\circ$. This set of four rotations defines the $p4$ group. 
A single base kernel is rotated by these four angles, producing four orientation-specific filters that capture features consistently across different orientations.

Let the four discrete rotations be indexed by \(r \in \{0,1,2,3\}\),
corresponding to angles \(0^\circ, 90^\circ, 180^\circ, 270^\circ\).
Let \(R_r\) denote the action that rotates spatial coordinates by the angle
associated with \(r\).
For an input
\(\mathbf{X} \in \mathbb{R}^{C_{\mathrm{in}} \times H \times W}\)
and kernels
\(\mathbf{W} \in \mathbb{R}^{C_{\mathrm{out}} \times C_{\mathrm{in}} \times K_h \times K_w}\),
the \(r\)-rotated kernels are defined by
\begin{equation}
\label{eq:p4-rot-kernel-op}
\mathbf{W}^{(r)}_{c_o,c_i}
\;=\;
R_r\!\big(\mathbf{W}_{c_o,c_i}\big),
\qquad
\substack{
\displaystyle c_o \in \{0,\dots,C_{\mathrm{out}}-1\},\\[2pt]
\displaystyle c_i \in \{0,\dots,C_{\mathrm{in}}-1\}.
}
\end{equation}
For valid convolution, the output spatial size is
\begin{equation}
\label{eq:p4-valid-sizes}
H' = H - K_h + 1,
\qquad
W' = W - K_w + 1.
\end{equation}
The output has an orientation channel for each \(r\):
\begin{equation}
\label{eq:p4-conv}
\begin{split}
Y_{c_o, r, h, w}
&=
\sum_{c_i=0}^{C_{\mathrm{in}}-1}
\sum_{i=0}^{K_h-1}
\sum_{j=0}^{K_w-1}
W^{(r)}_{c_o,c_i,i,j}\;
X_{c_i,\,h+i,\,w+j},
\\[6pt]
&\qquad
\substack{
\displaystyle c_o \in \{0,\dots,C_{\mathrm{out}}-1\},\\
\displaystyle r \in \{0,1,2,3\},\\
\displaystyle h \in \{0,\dots,H'-1\},\\
\displaystyle w \in \{0,\dots,W'-1\}
}
\end{split}
\end{equation}

\begin{figure}[h]
\centering
\includegraphics[width=12cm, angle=-90]{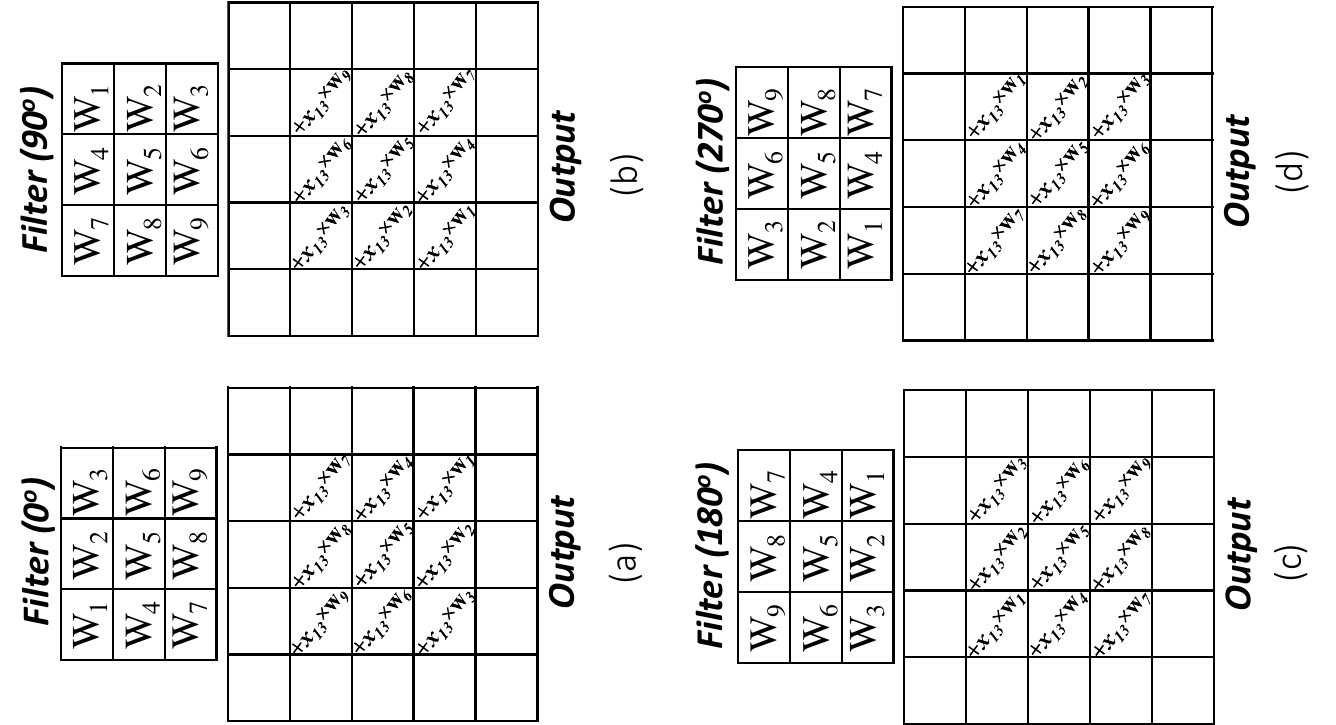}
\caption{Implementation of $p4$ group convolution with scatter operation where the original filter is rotated (a) $0^\circ$  (b) $90^\circ$  (c) $180^\circ$ (d) $270^\circ$.}
\label{fig:p4}
\end{figure}

While this improves rotational equivariance, it also increases computational cost because each rotated filter must be evaluated separately.
To mitigate this cost, we demonstrate how $p4$ group convolution on a 2D input can be accelerated using the scatter operation. In the $p4$ group, the original filter is rotated four times at $0^\circ$, $90^\circ$, $180^\circ$, and $270^\circ$, respectively. Figure~\ref{fig:p4} illustrates the scatter-based convolution process for each rotated filter when applied to the input pixel $x_{13}$.
It is clear the same product between a filter weight and an input element appears in all 4 convolutions albeit in different locations. For example, the multiplication $x_{13}\times w_9$ appears in 4 different corners of all convolutions. We can compute the multiplication $x_{13}\times w_9$ for the convolution with the first filter and use the result of this multiplication in the other convolutions directly. The multiplication is performed for only one filter. The index of the output location in which we add the result of the multiplication can be pre-computed for each filter. The same idea also can be applied to $p4m$ group convolution in which the original filter transformed 8 times (rotation and reflection). The multiplication is performed for the first filter and the result is reused by the other 7 filters directly.

Formally,
for the four symmetric rotations \(r \in \{0,1,2,3\}\), corresponding to 
\(0^\circ, 90^\circ, 180^\circ,\) and \(270^\circ\), let 
\((m',n') = R_r(m,n)\) denote the rotated kernel coordinates obtained by 
applying the spatial rotation operator \(R_r\) to the original offsets 
\((m,n)\). Under this rotation, the scatter-based convolution becomes
\begin{equation}
\label{eq:rot-scatter}
\begin{split}
Y_{c_o,\; i - m' + \lfloor K_h/2\rfloor,\;
               j - n' + \lfloor K_w/2\rfloor}
\; +=\; \\
\sum_{c_i=0}^{C_{\mathrm{in}}-1}
X_{c_i,h,w}\; W_{c_o,c_i,m,n},
\\[6pt]
\substack{
\displaystyle m \in \{0,\dots,K_h-1\},\quad
\displaystyle n \in \{0,\dots,K_w-1\},\\[4pt]
\displaystyle h \in \{0,\dots,H'-1\},\quad
\displaystyle w \in \{0,\dots,W'-1\},\\[4pt]
\displaystyle c_o \in \{0,\dots,C_{\mathrm{out}}-1\},\quad
\displaystyle r \in \{0,1,2,3\}.
}
\end{split}
\end{equation}
In this expression, the channel-wise multiplication and accumulation term 
\(\sum\limits_{c_i=0}^{C_{\mathrm{in}}-1} X_{c_i,h,w}\, W_{c_o,c_i,m,n}\) is computed only once. 
For each rotation \(r\), this accumulated product is added to a different output location 
because the kernel coordinate \((m,n)\) is transformed to its rotated counterpart 
\((m',n') = R_r(m,n)\). Thus, only the scatter location changes across rotations, 
while the channel-wise multiplication and accumulation is reused for all four orientations.

A rotation-invariant response can be obtained by group pooling over the four
orientations. For example, average pooling yields
\begin{equation}
\label{eq:p4-pooling-avg}
\widetilde{Y}_{c_o,h,w}
=
\frac{1}{4}\sum_{r=0}^{3} Y_{c_o,r,h,w}.
\end{equation}
Alternatively, max pooling over orientations may be used:
\begin{equation}
\label{eq:p4-pooling-max}
\widetilde{Y}_{c_o,h,w}
=
\max_{r \in \{0,1,2,3\}} \; Y_{c_o,r,h,w}.
\end{equation}

For symmetric rotation equivariant convolution, the number of filters will increase with the increase of the dimension of the input data. Consequently, the computational overhead also increases. However, with the scatter operation, the multiplication just needs to be performed for one filter and the result of the multiplication can be reused by the other filters directly. It means the number of multiplications is a constant (with respect to the number of directions) and the convolution is only dominated by the addition operation. 

Now we introduce the gradient computation for the rotation-invariant convolution.
During backpropagation, the loss gradient propagates from the output toward the input.
The incoming (upstream) gradient from the next layer is denoted by
\[
G_{c_o,h,w}
=
\frac{\partial\mathcal{L}}{\partial \widetilde{Y}_{c_o,h,w}}.
\]
where \(\widetilde{\mathbf{Y}}\) is the rotation-invariant output feature map obtained after
orientation pooling. 
We derive the gradients with respect to the
rotation-equivariant feature maps \(\mathbf{Y}_{c_o,r,h,w}\),
the input feature maps \(\mathbf{X}\),
and the base convolution kernel \(\mathbf{W}\).

For average pooling, the gradient is distributed equally:
\begin{equation}
\label{eq:rot-back-avg}
\frac{\partial\mathcal{L}}{\partial Y_{c_o,r,h,w}}
=
\frac{1}{4}\,G_{c_o,h,w}.
\end{equation}
For max pooling, we define
\[
r^\star(c_o,h,w)
\in
\arg\max_{r} Y_{c_o,r,h,w}.
\]
Then the gradient passes only through the maximal rotation:
\begin{equation}
\label{eq:rot-back-max}
\frac{\partial\mathcal{L}}{\partial Y_{c_o,r,h,w}}
=
G_{c_o,h,w}\;
\mathbf{1}\!\left[r = r^\star(c_o,h,w)\right].
\end{equation}
Let
\[
\Delta_{c_o,r,h,w}
=
\frac{\partial\mathcal{L}}{\partial Y_{c_o,r,h,w}}
\]
for compactness in the following derivations.

For each rotation \(r \in \{0,1,2,3\}\), the gradients with respect to the
input and the \(r\)-rotated kernel \(\mathbf{W}^{(r)}\)
are given by the standard convolution backward formulas:

\begin{equation}
\label{eq:rot-back-X-explicit}
\frac{\partial\mathcal{L}}{\partial X_{c_i,h,w}}^{(r)}
=
\sum_{c_o=0}^{C_{\mathrm{out}}-1}
\sum_{i=0}^{K_h-1}
\sum_{j=0}^{K_w-1}
W^{(r)}_{c_o,c_i,i,j}\;
\Delta_{c_o,r,\,h-i,\,w-j},
\end{equation}

\begin{equation}
\label{eq:rot-back-W-explicit}
\frac{\partial\mathcal{L}}{\partial W^{(r)}_{c_o,c_i,i,j}}
=
\sum_{h=0}^{H'-1}
\sum_{w=0}^{W'-1}
X_{c_i,\,h+i,\,w+j}\;
\Delta_{c_o,r,h,w}.
\end{equation}

The total gradient with respect to the input feature map is the sum over
the four rotated branches:
\begin{equation}
\label{eq:rot-back-X-total}
\frac{\partial\mathcal{L}}{\partial X_{c_i,h,w}}
=
\sum_{r=0}^{3}
\frac{\partial\mathcal{L}}{\partial X_{c_i,h,w}}^{(r)}.
\end{equation}

To obtain the gradient with respect to the unrotated base kernel
\(\mathbf{W}\),
each rotated kernel gradient is first inverse-rotated
and then accumulated:
\begin{equation}
\label{eq:rot-back-W-total}
\frac{\partial\mathcal{L}}{\partial W_{c_o,c_i,i,j}}
=
\sum_{r=0}^{3}
R_{-r}\!\left(
\frac{\partial\mathcal{L}}{\partial W^{(r)}_{c_o,c_i,i,j}}
\right).
\end{equation}
Here \(R_{-r}\) denotes a rotation of the gradient by \(-r\times90^\circ\),
mapping each rotated kernel’s gradient back to the base orientation.
The proposed scatter approach can also accelerate backpropagation, since the gradient computations themselves are convolution operations, as shown in these formulations.

\subsection{Parallelization of the scatter operation} \label{sec:para}

\begin{figure}[h]
\centering
\includegraphics[width=8cm]{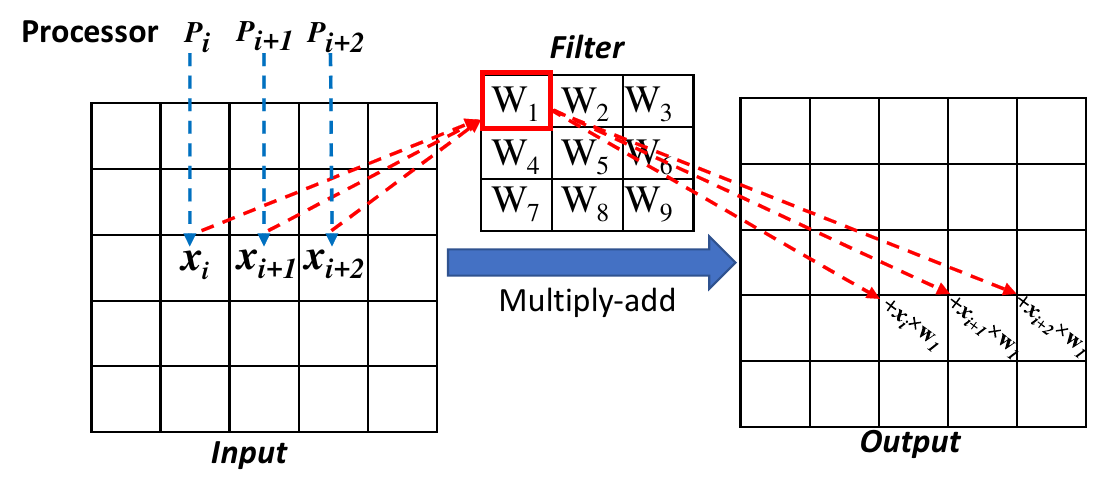}
\caption{Parallel scatter operation, where each processor $P_i$ corresponds to each input data.}
\label{fig:PSO}
\end{figure}

The parallelization of the convolution with scatter operation is shown in Fig. \ref{fig:PSO}.  In each parallel step, every processor $P_i$ reads the corresponding input data $x_i$ and multiplies it with the same filter weight, then the results of those multiplications will be added to the different output locations. Multiplication with the same filter weight can avoid Write after Write hazard, if there is a synchronization between two parallel steps. For G-convolution, the multiplication is performed only once using this approach.

\begin{figure}[h]
\centering
\includegraphics[width=8cm]{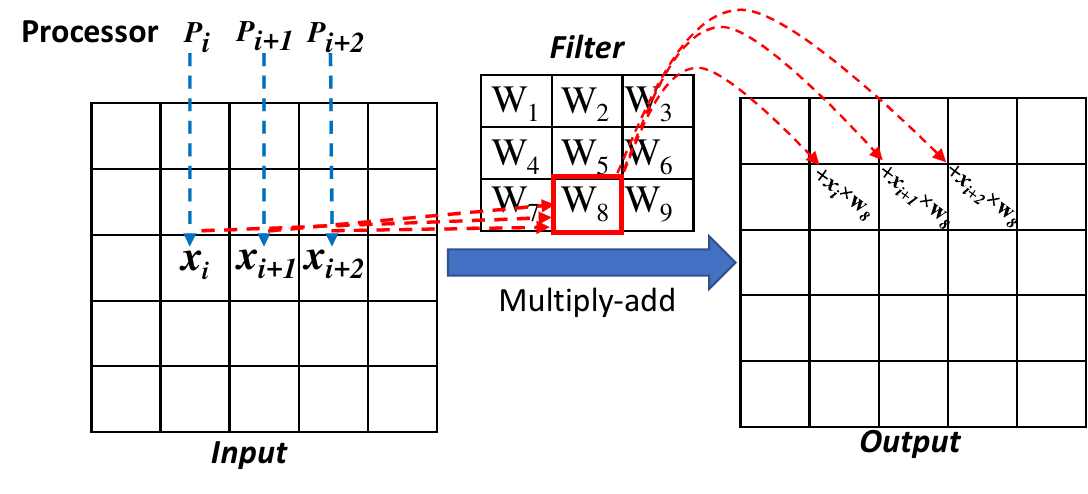}
\caption{Illustration of scatter operation by multiplying each input data with another filter weight. No matter which filter weight it is multiplied by, all writing positions in the output are always in the same row.}
\label{fig:cache-cohe}
\end{figure}

\begin{figure}[h]
\centering
\includegraphics[width=8cm]{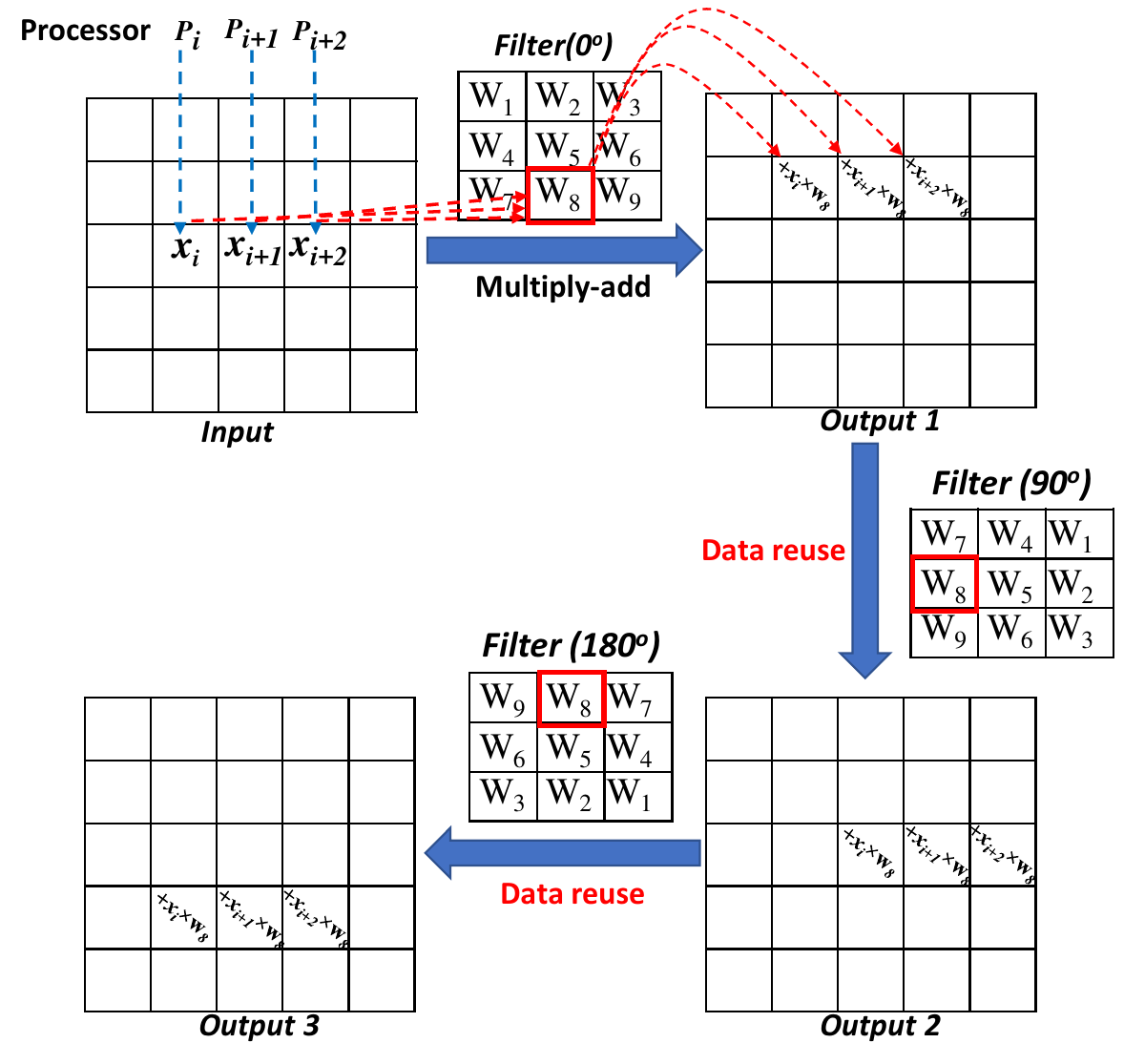}
\caption{Output positions for the rotated filter with data reuse. No matter how the filter is rotated, all writing positions in the output are always in the same row if the processors process input data in one row.}
\label{fig:cache-cohe-rot}
\end{figure}

The scatter operation is a cache friendly operation and can be easily vectorized. As shown in Fig. \ref{fig:cache-cohe}, if all reading positions in the input data are in the same row and are multiplied by the same filter weight, then all writing positions in the output are always in the same row, no matter which filter weight is  multiplied. Even if the filter is rotated symmetrically, all writing positions in the output are also in the same row, see Fig. \ref{fig:cache-cohe-rot}. It meets the data locality requirement of modern SIMD architecture. On the other hand, this figure also shows us a way to reduce the number of synchronizations required. All processors multiply its own input data with the same filter weight and add the result in 4 different outputs. Obviously, before completing the last addition, all processors have no need to be synchronized since they are writing different memory positions of different outputs. We just insert a synchronization after the last addition. The number of synchronizations can be reduced 4 times. 
 
\subsection{Arbitrary rotation convolution formulation} \label{sec:fft_arbitrary}
An interpolation is required for sampling after rotation with arbitrary angle. 
However, it can introduce artifacts that degrade the performance of rotation-invariant models.
In this subsection, we propose an approach which takes the advantage of symmetric rotation and steerable filter to implement rotation-invariant convolution with filter arbitrarily rotated. 
The proposed method accelerates the training process while eliminating the need for interpolation.
\begin{figure}[ht]
\begin{center}
\includegraphics[scale=0.44]{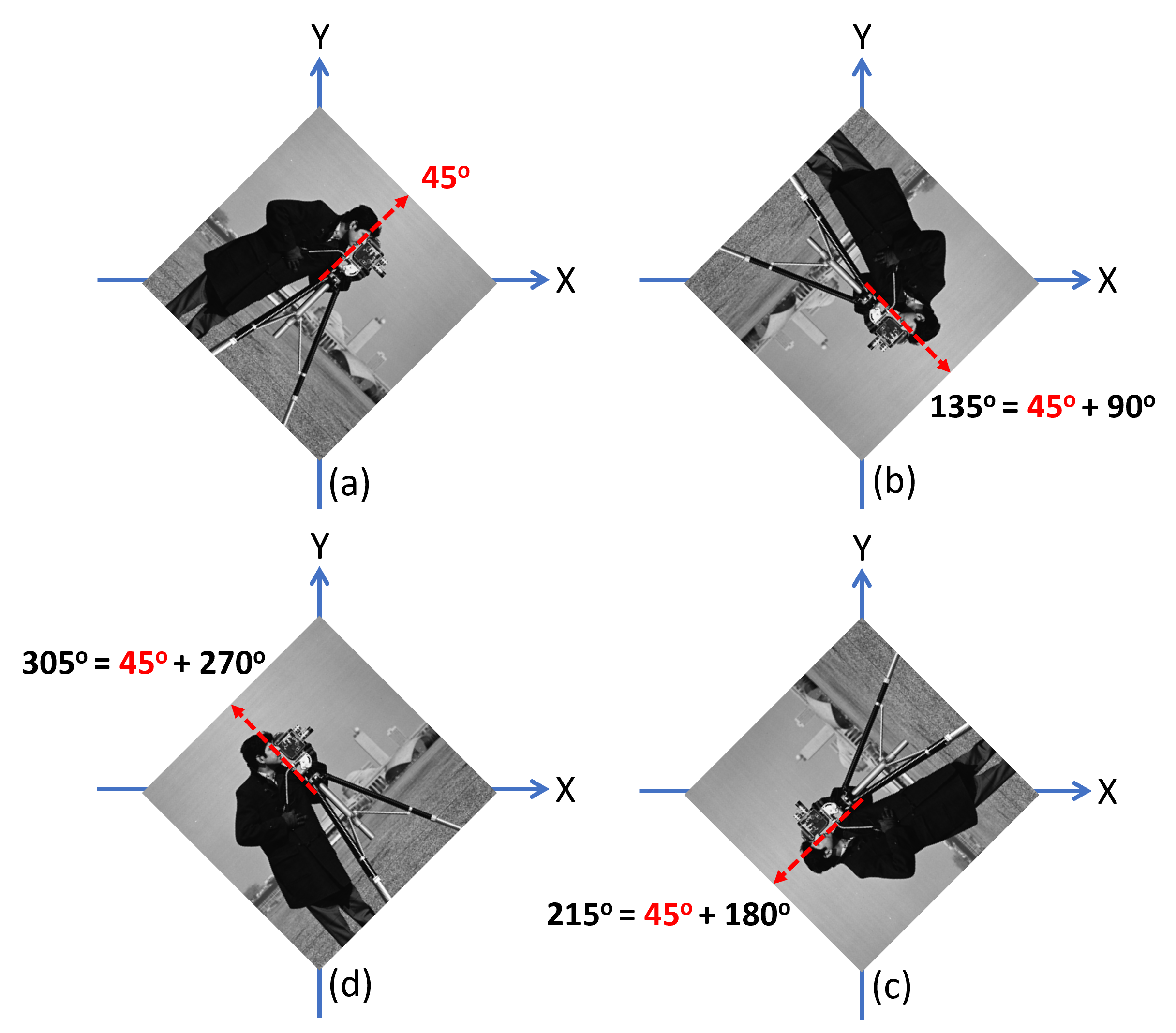}
\caption{Symmetric correspondents of $45^\circ$ rotated image in other quadrants of coordinate plane}
\label{fig:quadrant}
\end{center}
\end{figure}

The sampled orientations are usually evenly distributed in rotation space for rotation equivariance convolution. For example, the angle increment between every two successive rotations is $\frac{\pi}{12}$, if the number of sampled orientations is 24. We observe the rotation angle in the first quadrant of coordinate plane always has a symmetric correspondent in each of other quadrants, see Fig. \ref{fig:quadrant}. As shown in the figure, the rotated versions of image with angles $135^\circ$, $215^\circ$ and $305^\circ$ can be obtained by symmetrically rotating the $45^\circ$-rotated image by $90^\circ$, $180^\circ$ and $270^\circ$, respectively.    
This observation gives us a chance to accelerate the rotation equivariance convolution with filter arbitrarily rotated. We apply filter rotations when the rotation angle falls within the first quadrant. Rotations in other quadrants can be obtained by symmetrically mapping the correspondence from the first quadrant. The number of rotations is reduced by 75$\%$. This indicate that we can rotate the filter in the first quadrant of coordinate plane, and perform the multiplication for the rotations in the first quadrant. We then reuse these multiplication results for the rotation in the other quadrants. 

As steerable filters can generate rotated responses without relying on interpolation,  
we construct the filter in the first quadrant of the coordinate plane using the following steerable filter formulation as in \cite{93808}:
\begin{equation}
\psi_\theta(x, y) = \sin(\theta) \cdot f_x(x, y) + \cos(\theta) \cdot f_y(x, y)
\label{eq:steerable_filter}
\end{equation}
Equation~\eqref{eq:steerable_filter} defines the steerable filter \( \psi_\theta(x, y) \) at a given rotation angle \( \theta \),  
where \( (x, y) \in \mathbb{R}^2 \) denotes the spatial coordinates.
The functions \( f_x(x, y) \) and \( f_y(x, y) \) represent the learned base filters oriented along the horizontal (x-axis)  
and vertical (y-axis) directions, respectively. The steerable filter \( \psi_\theta(x, y) \) is constructed as a linear combination  
of these directional basis filters using fixed orientation-dependent coefficients \( \sin(\theta) \) and \( \cos(\theta) \),  
enabling efficient and continuous steering of the filter to any desired angle \( \theta \).

To encourage the learned base filters to exhibit steerable behavior, we add two regularization losses to the standard
cross-entropy objective. The overall loss is defined as:
\begin{equation}
\mathcal{L}_{\text{total}}
=
\mathcal{L}_{\text{CE}}
+
\lambda_{\text{mag}}\,\mathcal{L}_{\text{mag}}
+
\lambda_{\text{orth}}\,\mathcal{L}_{\text{orth}} ,
\label{eq:total_loss}
\end{equation}
where \( \mathcal{L}_{\text{CE}} \) denotes the task-specific cross-entropy loss,
and \( \lambda_{\text{mag}} \) and \( \lambda_{\text{orth}} \) are scalar weighting coefficients controlling
the strength of the magnitude and orthogonality regularizations, respectively.

The magnitude-matching loss encourages the two base filters to maintain similar energy levels:
\begin{equation}
\mathcal{L}_{\text{mag}}
=
\frac{1}{B}
\sum_{b=1}^{B}
\left(
\left\|\mathbf{w}^{(x)}_{b}\right\|_{2}
-
\left\|\mathbf{w}^{(y)}_{b}\right\|_{2}
\right)^{2},
\label{eq:mag_loss}
\end{equation}
where \( \mathbf{w}^{(x)}_{b} \) and \( \mathbf{w}^{(y)}_{b} \) denote the flattened kernel weights associated with 
the \( b \)-th filter in the base filters \( f_x \) and \( f_y \), respectively, and \( B \) is the total number of filters.

The orthogonality loss penalizes the correlation between the two directional bases, promoting rotational independence:
\begin{equation}
\mathcal{L}_{\text{orth}}
=
\frac{1}{B}
\sum_{b=1}^{B}
\left(
\frac{
\left\langle \mathbf{w}^{(x)}_{b}, \mathbf{w}^{(y)}_{b} \right\rangle
}{
\left\|\mathbf{w}^{(x)}_{b}\right\|_{2}
\left\|\mathbf{w}^{(y)}_{b}\right\|_{2}
+ \varepsilon
}
\right)^{2},
\label{eq:orth_loss}
\end{equation}
where \( \varepsilon \!>\! 0 \) ensures numerical stability during training.
Together, these regularization terms guide the base filters \( f_x \) and \( f_y \)
toward forming an approximately steerable pair that preserves orientation consistency across rotations.

\section{GPU implementation}\label{sec:GPU}

\subsection{GPU implementation of single-orientation convolution with scatter operation}\label{sec:GPU}
The convolution operation can be efficiently implemented on GPU using matrix multiplication, as shown in \cite{cudnn-arxiv}. This approach requires transforming the input image and the convolution kernel into matrices that can exploit the fast multiplication capabilities of the GPU. This transformation involves duplicating the input data to lower its dimensionality. 

In our GPU implementation of convolution with scatter operation, we use matrix multiplication to perform the channel-wise multiplication and summation of the input and a filter kernel. To achieve a higher performance,
NVIDIA cuBLAS library  \cite{cublas}, a library of highly optimized matrix multiplication routines, is used to perform the channel-wise multiplication and summation. The results are then scattered to the corresponding output positions according to the scattering rule. Unlike the method in \cite{cudnn-arxiv}, we do not apply any data lowering in our GPU implementation. Our GPU implementation follows gather-GEMM-scatter dataflow \cite{torchsparse} but gather without data lowering. 

\begin{figure}[h]
\centering
    \includegraphics[width=8.5cm]{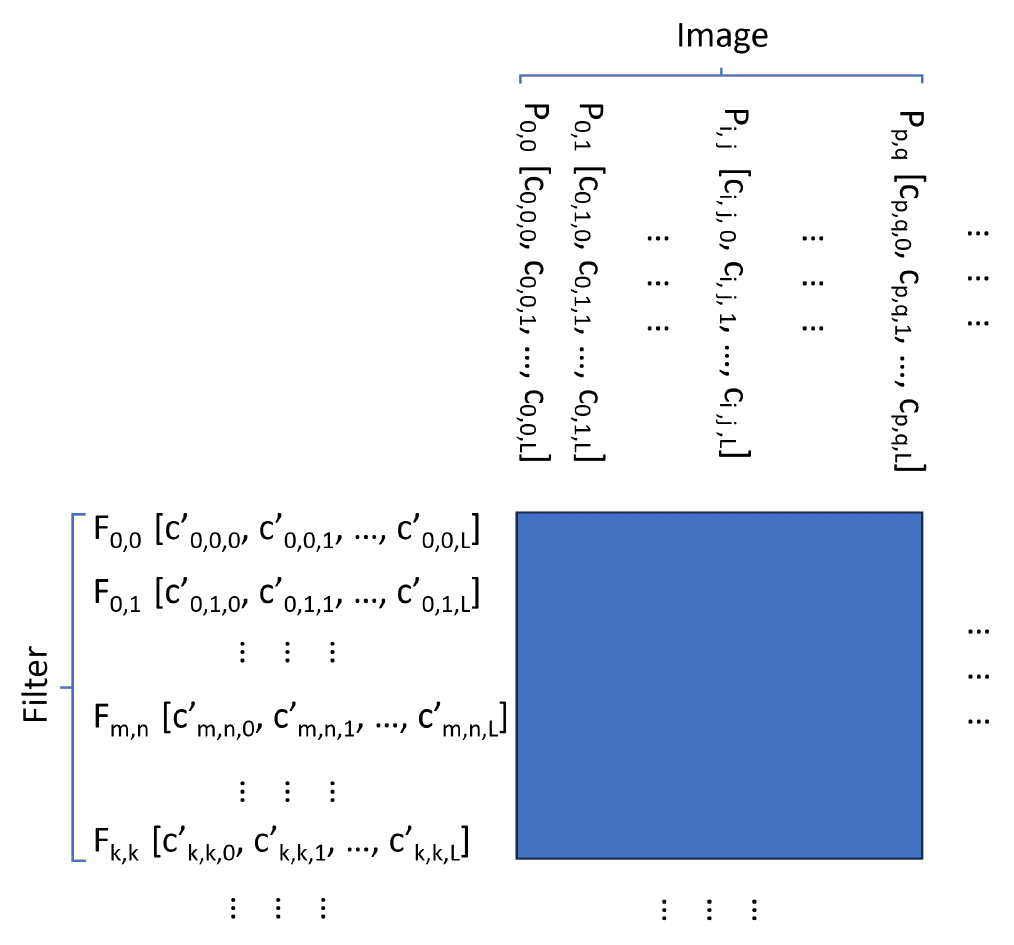}
\caption{Memory layout of the input image and filter kernel for matrix multiplication, where $P_{i,j}$ represents an image pixel,  $F_{m,n}$ represents an element of the filter kernel, $c_{i,j,L}$ and $c'_{m,n,L}$ are channel values. Only a single image and a single filter are depicted.} 
\label{fig:mem-layout}
\end{figure}

Considering that cuBLAS accesses matrices in column-major order, we arrange the input image in CNHW layout and the filter in NHWC layout, as shown in Fig. \ref{fig:mem-layout}. The resulting matrix ensures coalesced access for the parallel implementation of the scatter operation, as described in Section \ref{sec:para}.

\begin{figure}[h]
\centering
    \includegraphics[width=5.5cm]{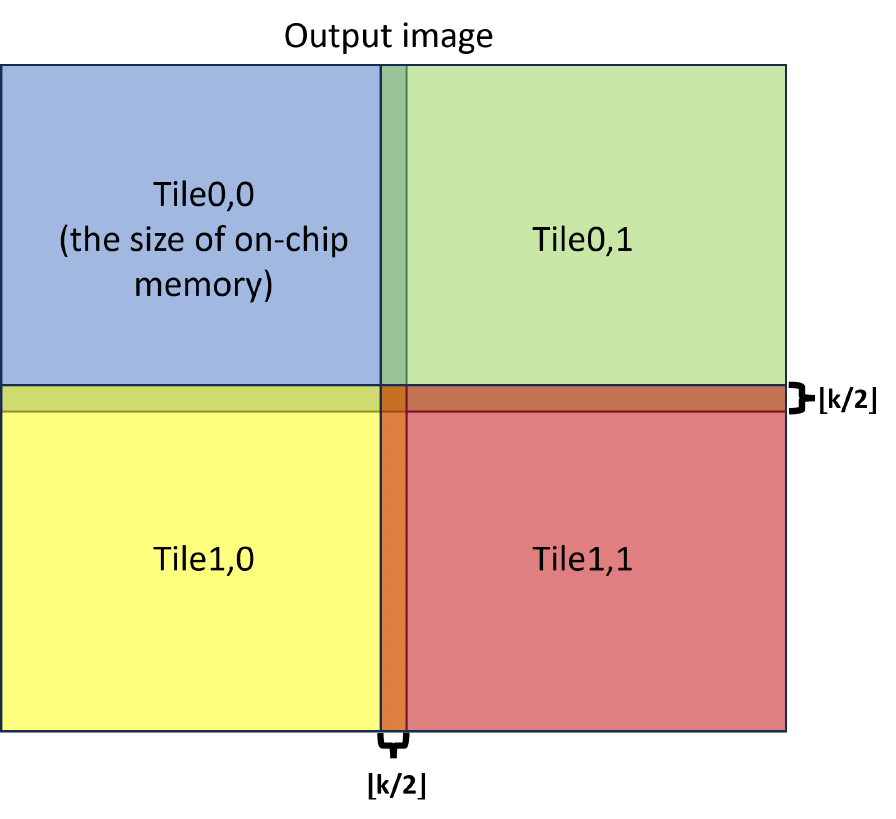}
\caption{Image tiling with on-chip memory of GPU, where the size of overlap between tiles is $\floor {k /2}$.} 
\vspace{-0.4cm}
\label{fig:imge-tile}
\end{figure}

After performing matrix multiplication, the results need to be added to the output image using the parallel scattering operation. However, directly writing to off-chip memory (i.e., the global memory of CUDA) is time-consuming, as each element of the output image needs to be accessed $k^2$ times. To avoid this, we tile the output image as in Fig. \ref{fig:imge-tile} so that each tile can be accommodated in on-chip memory (i.e., the shared memory of CUDA). The scatter operation will be performed in on-chip memory, and the results will then be written to off-chip memory. There is an overlapping region between adjacent tiles, which ensures that elements on the borders can be scattered to neighboring positions completely.

\subsection{GPU implementation of symmetric rotation invariant convolution with scatter operation}\label{sec:GPU}
The main idea behind accelerating symmetric rotation invariant convolution using the scatter operation is the reuse of intermediate multiplication results, as described in Section \ref{subsec:accSymRot}. The matrix-multiplication outputs are shared across all rotations, eliminating redundant computation.
To make the layer rotation invariant, an orientation-pooling operation is applied afterward, which can be implemented in two ways: point-wise average pooling or point-wise max pooling.

For point-wise average pooling, each element in the result of the matrix multiplication is divided by $r$, the number of rotations, before being added to the output image. This means that the required on-chip memory size is equal to the tile size. However, for point-wise max pooling, the on-chip memory must be twice the tile size—one part for computing the current tile and another for storing the previous maximum values. After finishing additions for the current tile, the point-wise max operation is performed between the current tile and the previously stored max tile. This is necessary because max pooling is a non-linear operation, meaning the maximum value cannot be determined before the additions are completed. Additionally, the overlapping region between adjacent tiles must be extended to $\floor {k /2} + 1$ to ensure correct max pooling calculations.  Only the central part of each tile, excluding $\floor {k /2} + 1$ border lines, is used in the final output. This ensures that the max values are computed correctly without missing contributions from neighboring tiles.

\section{Performance Evaluation}

We implemented the rotation-invariant convolution using four different approaches: (1) our scatter based kernel, (2) a cuDNN based pipeline, (3) an Oriented Response Networks (ORN) implementation, and (4) an equivariant convolution implementation using the e2cnn library. 

In our scatter based approach, the four symmetric rotations are processed within a single fused convolution, so each input element is multiplied only once.
In contrast, the cuDNN based pipeline must treat every rotated filter as an independent kernel, performing four separate convolution and therefore four distinct multiply–accumulate passes before the results can be combined. 
The ORN and e2cnn implementations follow their respective rotation-equivariant formulations and are included as additional baselines for comparison.
Among these, we implemented the backward pass for the scatter-based and cuDNN-based versions, as automatic gradient computation is not available for these custom rotation-invariant operations, ensuring that gradients propagate correctly through every rotated filter.

\subsection{Model for Evaluation}\label{subsubsec:model}
\begin{figure}[h]
    \centering
    \begin{subfigure}[b]{0.49\textwidth}  
        \centering
        \includegraphics[width=\textwidth]{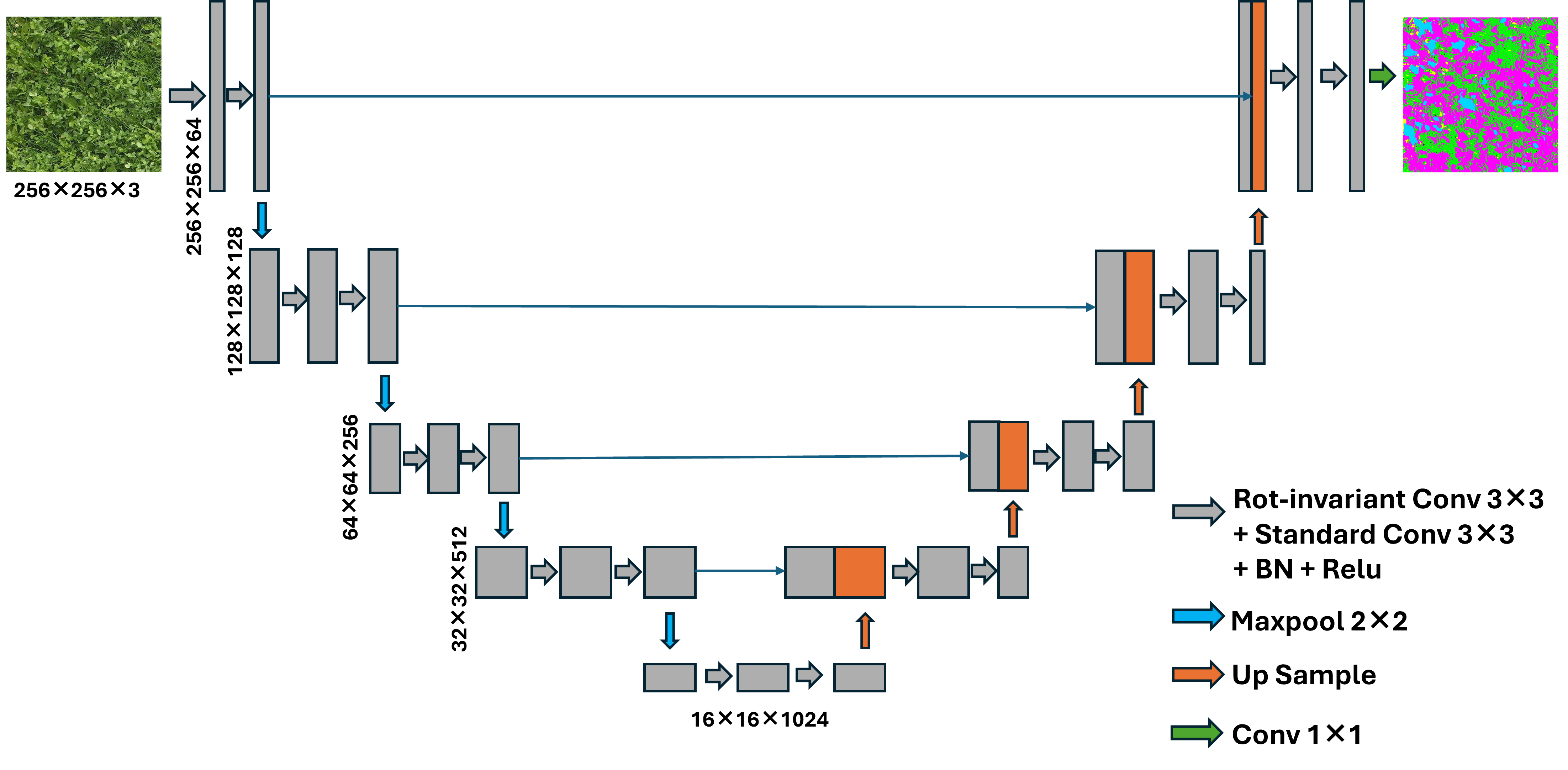}
    \end{subfigure}
    \caption{Unet architecture with rotation invariant convolution block.}
	\vspace{-0.4cm}
    \label{fig:unet}
\end{figure}

To evaluate the performance of the scatter based rotation invariant convolution in a real segmentation application, we replaced the first convolution of base convolution block in UNet \cite{ronneberger2015u} with the rotation invariant convolution described above (see Fig. \ref{fig:unet} for illustration). To balance performance and dimension explosion, we apply max pooling across every four symmetric rotations; for example, with eight rotations, this produces two orientation features, each corresponding to a group of four symmetric rotations. The second convolution within the base block uses a standard convolution, we found through experiments that this design improves performance.

\subsection{Datasets for Training and Testing}\label{subsec:dataset}
The modified UNet were trained and tested on two datasets: an in-farm plant-segmentation dataset captured with a DJI M300 drone \cite{dji_m300} equipped with a P1 photogrammetry camera \cite{dji_p1}, and the public Semantic Segmentation Drone Dataset \cite{semantic_drone_dataset}. All experiments were conducted on an Ubuntu 22.04 system equipped with eight NVIDIA RTX 3090 GPUs.

\subsubsection{Plant segmentation dataset}\label{subsubsec:plant}
Accurately measuring species composition in pasture and forage systems is fundamental for biodiversity monitoring and precision agriculture because it informs management decisions on seeding, fertilization, and grazing intensity. However, mixed‑species pastures are highly heterogeneous: the realized stand rarely matches the seed plan and can vary widely across a field and over time. This variability places stringent demands on automated, high‑resolution, plant‑level segmentation of aerial/UAV imagery. Because leaves and other canopy structures appear at arbitrary orientations in overhead views, incorporating rotation invariance into the segmentation pipeline is critical for robust species discrimination in multi‑species pastures.

We mapped pasture species composition from high‑resolution UAV imagery acquired with a DJI drone equipped with a Zenmuse P1 camera. Flights were conducted at approximately 12 m above ground level,
yielding a nominal ground sampling distance (GSD) of {0.15} cm per pixel.
Forward image overlap was 80\%,
and each P1 frame covered roughly {12.31 x 8.20} m$^2$ on the ground.
This dataset contains 1000 samples and we use 800 samples for training and 100 for validation and 100 for test.

\subsubsection{Semantic Segmentation Drone dataset}\label{subsec:droneDataset}
The Semantic Drone Segmentation dataset contains 400 high-resolution nadir-view images of urban residential scenes, captured at altitudes between 5 m and 30 m above ground and annotated with detailed pixel-level semantic labels covering 24 object categories. In our experiments, we use 340 images for training, 32 for validation, and 28 for testing.

\subsection{Evaluation with Different Convolution Approaches}\label{subsubsec:differentConvs}

\subsubsection{Single orientation rotation convolution}\label{subsubsec:singleOri}
Given that single-orientation rotation convolution collapses to standard convolution, we first conduct a direct comparison between our scatter-based convolution API and cuDNN’s standard convolution API. The corresponding results are reported in Appendix~\ref{appendix:compareCUDNN}.

\begin{figure}[htbp]
  \centering

  \begin{subfigure}{0.45\textwidth}
    \centering
    \includegraphics[width=\textwidth]{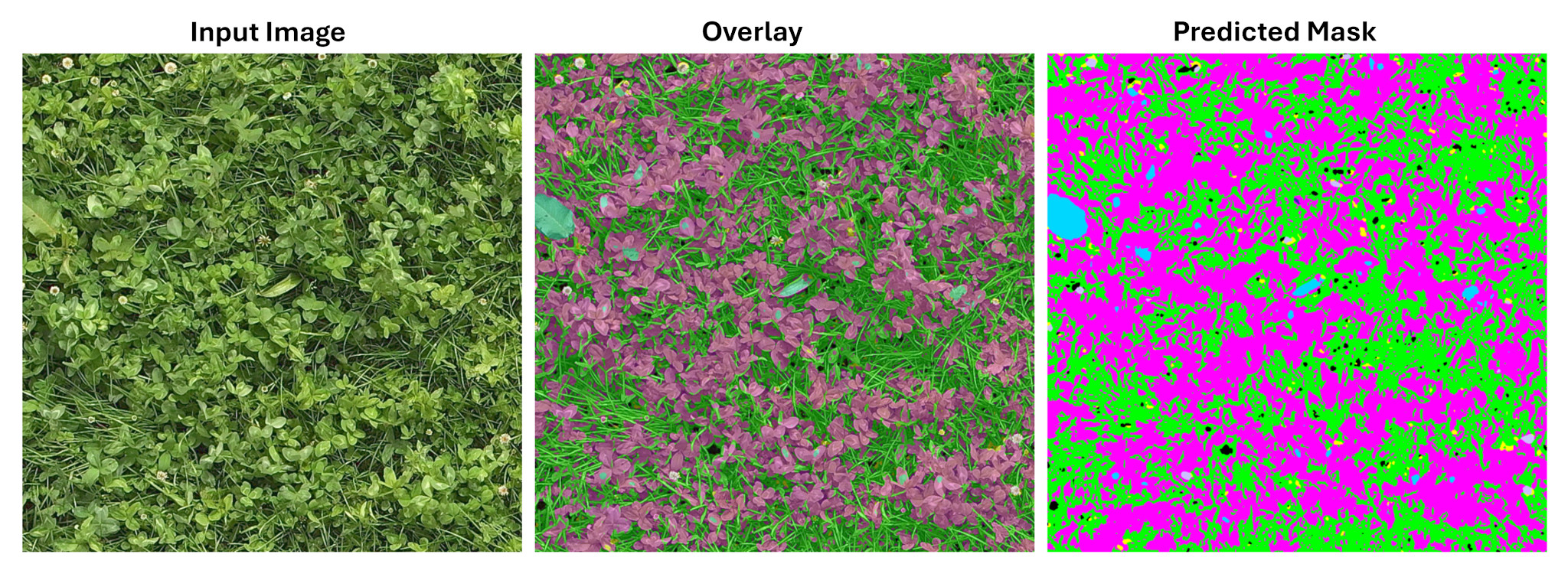}
  \end{subfigure}
  \par\vspace{0.3em}

  \begin{subfigure}{0.45\textwidth}
    \centering
    \includegraphics[width=\textwidth]{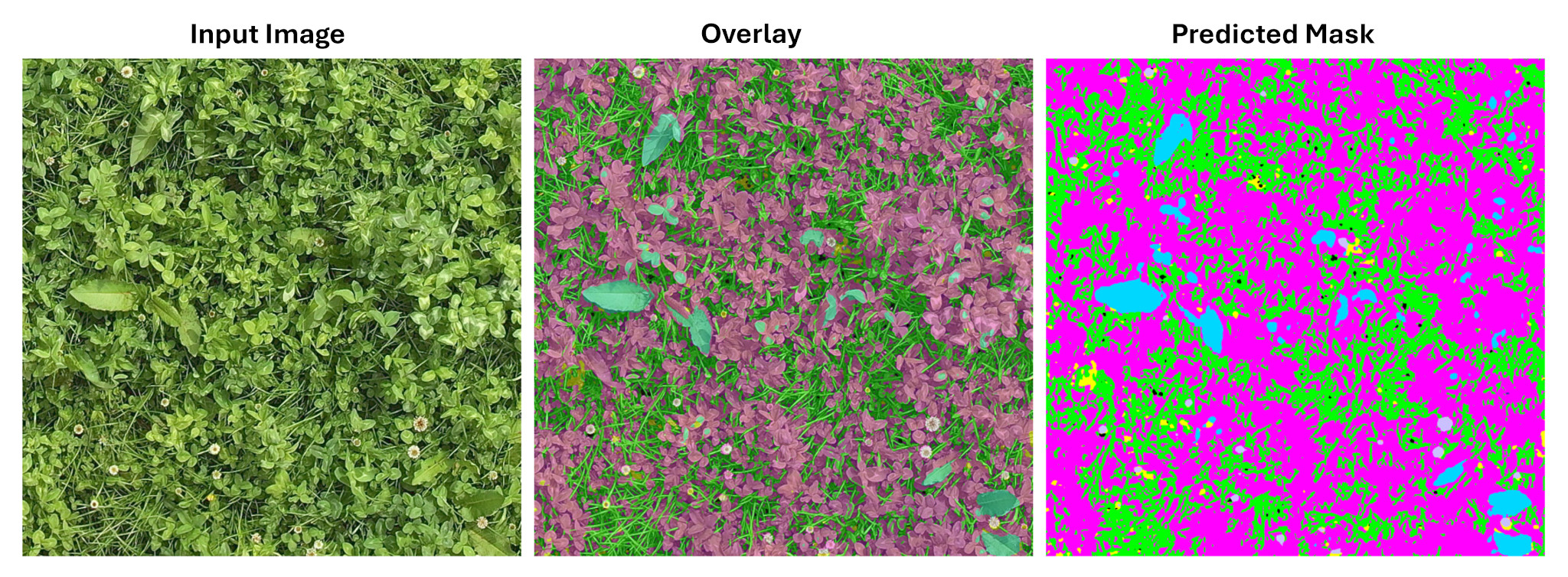}
  \end{subfigure}
  \par\vspace{0.3em}

  \begin{subfigure}{0.45\textwidth}
    \centering
    \includegraphics[width=\textwidth]{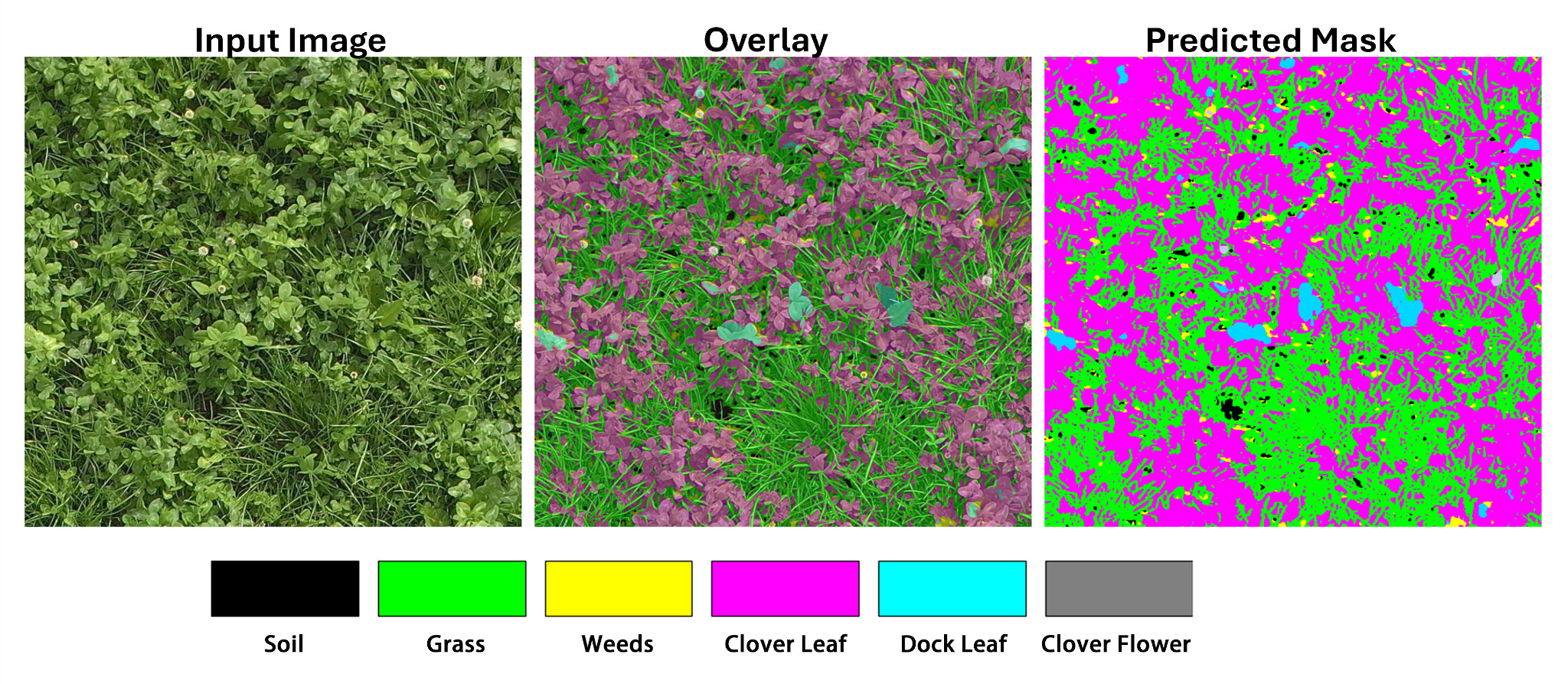}
  \end{subfigure}

  \caption{Visualization of segmentation results on plant data.}
  \label{fig:plant}
\end{figure}

\begin{figure}[htbp]
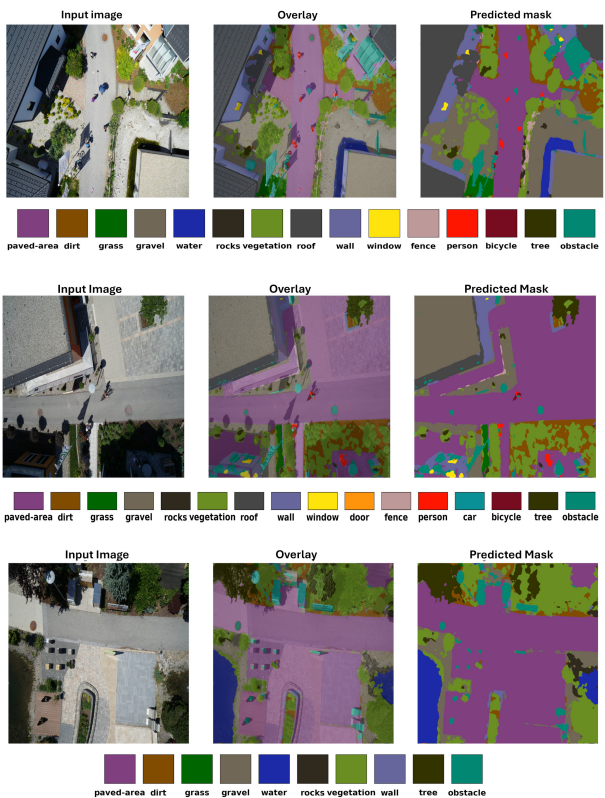

  \centering
  
  \begin{subfigure}{0.45\textwidth}
    \centering
    \includegraphics[width=\textwidth]{fig/FL\_timing/infer-img-drone/drone-seg-1.jpg}
   
  \end{subfigure}\par\medskip
  
  \begin{subfigure}{0.45\textwidth}
    \centering
    \includegraphics[width=\textwidth]{fig/FL\_timing/infer-img-drone/drone-seg-2.jpg}
  
  \end{subfigure}\par\medskip
  
  \begin{subfigure}{0.44\textwidth}
    \centering
    \includegraphics[width=\textwidth]{fig/FL\_timing/infer-img-drone/drone-seg-3.jpg}
  
  \end{subfigure}

  \caption{Visualization of segmentation results on Semantic Segmentation Drone dataset.}
  \label{fig:drone}
\end{figure}

We compare our approach with cuDNN as well as the state-of-the-art ORN \cite{orn} and E2CNN \cite{e2cnn} frameworks, both of which provide APIs for achieving rotation invariance.
Tables \ref{tab:single-plant} and \ref{tab:single-drone} report the performance of cuDNN, ORN, E2CNN, and the proposed scatter-based convolution under the single-orientation setting.
The observed accuracy variation is within $\pm 0.1\%$.
Since single-orientation rotation convolution degenerates to standard convolution, all methods achieve nearly identical segmentation accuracy. Nevertheless, notable differences arise in computational efficiency.
On the plant dataset with 256×256 inputs, E2CNN achieves the lowest training time and energy consumption, while scatter-based convolution also demonstrates competitive efficiency, training approximately 9.2\% faster than cuDNN and reducing energy consumption by about 6.9\%. At 1024×1024 resolution, scatter-based convolution achieves the best efficiency, reducing training time by roughly 7.0\% and energy usage by 7.6\% compared with cuDNN.
Similar trends are observed on the drone dataset: at 256 resolution, E2CNN attains the highest accuracy and lowest computational cost, whereas scatter-based convolution remains consistently more efficient than cuDNN. At 1024 resolution, scatter-based convolution outperforms all baselines in both training time (7.1\% reduction) and energy consumption (7.9\% reduction).
These results indicate that while E2CNN is particularly efficient for smaller input resolutions, the proposed scatter-based convolution provides consistent and scalable efficiency gains—especially at higher resolutions—without sacrificing segmentation accuracy, as it avoids explicit data lowering and reduces memory traffic through direct scatter accumulation.
Representative segmentation results are visulized in Fig. \ref{fig:plant} and Fig. \ref{fig:drone}.


\begin{table}[ht]
\centering
\caption{Plant Data Segmentation Using Various Single-Orientation Rotation-Invariant Convolutions}
\renewcommand{\arraystretch}{1.2}
\begin{tabular}{c|c|c|c|c} 
\hline\hline
\shortstack{Image\\size} & 
\shortstack{Convolution\\applied} & 
\shortstack{Training\\time (s)} &  
\shortstack{Energy\\(kWh)} &  
\shortstack{Test\\accuracy (\%)} \\ 
\hline\hline
\multirow{4}{*}{256} 
 & cudnn based conv    & 8604.0  & 0.379  &73.59  \\ \cline{2-5}
 & ORN conv            & 7758.0  & 0.362  &73.63  \\ \cline{2-5}
 & E2CNN conv          & 7116.0  & 0.315  & 73.73 \\ \cline{2-5}
 & scatter based conv  & 7808.0  & 0.353  &73.62  \\
\hline

\multirow{4}{*}{1024} 
 & cudnn based conv    & 26153.0 & 1.758  &71.32  \\ \cline{2-5}
 & ORN conv            & 25538.0 & 1.719  &71.38  \\ \cline{2-5}
 & E2CNN conv          & 27405.0 & 2.213  &71.66  \\ \cline{2-5}
 & scatter based conv  & 24317.0 & 1.625  &71.28  \\
\hline
\end{tabular}
\label{tab:single-plant}
\end{table}

\begin{table}[ht]
\centering
\caption{Semantic drone segmentation Using Various Single-Orientation Rotation-Invariant Convolutions}
\renewcommand{\arraystretch}{1.2}
\begin{tabular}{c|c|c|c|c} 
\hline\hline
\shortstack{Image\\size} & 
\shortstack{Convolution\\applied} & 
\shortstack{Training\\time (s)} &  
\shortstack{Energy\\(kWh)} &  
\shortstack{Test\\accuracy (\%)} \\ 
\hline\hline
\multirow{4}{*}{256} 
 & cudnn based conv    & 3242.2  & 0.1408 &90.62  \\ \cline{2-5}
 & ORN conv    & 2945.1 & 0.1359  &90.67  \\ \cline{2-5}
 & E2CNN conv    &2705.7 &0.1186  &90.73  \\ \cline{2-5}
 & scatter based conv  & 2925.6  & 0.1325 &90.65  \\
\hline

\multirow{4}{*}{1024} 
  & cudnn based conv   & 9865.5  &0.6618 &88.71  \\ \cline{2-5}
 & ORN conv    &9726.3  &0.6498  &88.74  \\ \cline{2-5}
 & E2CNN conv    &10392.7 & 0.8372  & 88.78 \\ \cline{2-5}
 & scatter based conv  &9166.9  & 0.6094&88.69  \\
\hline
\end{tabular}
\label{tab:single-drone}
\end{table}

\subsubsection{Symetric four orientation rotation convolution}\label{subsubsec:fourOri}

Tables \ref{tab:four-plant} and \ref{tab:four-drone} summarize the results for the symmetric four-orientation configuration. Relative to single-orientation convolution, accuracy increases by about 2\%, with E2CNN achieving the highest accuracy, reflecting the benefit of four rotation modeling. The computational advantages of scatter-based convolution become more pronounced under this setting. On the plant dataset, scatter reduces training time by approximately 45.9\% at 256 resolution and 28.8\% at 1024 resolution, while lowering energy consumption by about 38.4\% and 18.6\%, respectively. On the drone dataset, scatter provides comparable benefits, achieving about 46.7\% faster training at 256 resolution and 32.9\% at 1024 resolution, with corresponding energy reductions of roughly 39.1\% and 22.8\%. These results indicate that symmetric four-orientation convolutions achieve strong segmentation accuracy with manageable computational cost. The scatter based convolution provides the most favorable accuracy–efficiency profile, benefiting from reuse of intermediate computations across symmetric rotations. This enables substantial reductions in training time and energy consumption while maintaining segmentation accuracy.

\begin{table}[ht]
\centering
\caption{Plant Data Segmentation Using Various Four-Orientation Rotation-Invariant Convolutions}
\renewcommand{\arraystretch}{1.2}
\begin{tabular}{c|c|c|c|c} 
\hline\hline
\shortstack{Image\\size} & 
\shortstack{Convolution\\applied} & 
\shortstack{Training\\time (s)} &  
\shortstack{Energy\\(kWh)} &  
\shortstack{Test\\accuracy (\%)} \\ 
\hline\hline
\multirow{4}{*}{256} 
 & cudnn based conv    & 20234.6  & 0.943  &75.64  \\ \cline{2-5}
 & ORN conv            & 20154.2  & 0.921  &  75.69\\ \cline{2-5}
 & E2CNN conv          & 18575.8  & 0.783  & 75.71 \\ \cline{2-5}
 & scatter based conv  & 10962.6  & 0.581  & 75.61 \\
\hline
\multirow{4}{*}{1024} 
 & cudnn based conv    & 82081.0  & 5.564   &72.92  \\ \cline{2-5}
 & ORN conv            & 81706.5  & 5.623   &72.98  \\ \cline{2-5}
 & E2CNN conv          & 76175.0  & 5.118   &73.05  \\ \cline{2-5}
 & scatter based conv  & 58433.9  & 4.532   &72.95  \\
\hline
\end{tabular}
\label{tab:four-plant}
\end{table}

\begin{table}[ht]
\centering
\caption{Semantic drone segmentation Using Various Four-Orientation Rotation-Invariant Convolutions}
\renewcommand{\arraystretch}{1.2}
\begin{tabular}{c|c|c|c|c} 
\hline\hline
\shortstack{Image\\size} & 
\shortstack{Convolution\\applied} & 
\shortstack{Training\\time (s)} &  
\shortstack{Energy\\(kWh)} &  
\shortstack{Test\\accuracy (\%)} \\ 
\hline\hline
\multirow{4}{*}{256} 
 & cudnn based conv    &7713.1  &0.3570 &91.88  \\ \cline{2-5}
 & ORN conv    &7651.6  &0.3501  &91.93  \\ \cline{2-5}
 & E2CNN conv    &6908.2 &0.2943  &92.06  \\ \cline{2-5}
 & scatter based conv  &4109.6  &0.2175 &91.88  \\
\hline
\multirow{4}{*}{1024} 
  & cudnn based conv    &30936.5  &2.103 &90.39  \\ \cline{2-5}
 & ORN conv    &31025.2  & 2.203  &90.41  \\ \cline{2-5}
 & E2CNN conv    &28925.6 &1.946  &90.62  \\ \cline{2-5}
 & scatter based conv  &20761.8  & 1.623&90.43  \\
\hline
\end{tabular}
\label{tab:four-drone}
\end{table}

\subsubsection{Symetric eight orientation rotation convolution}\label{subsubsec:eightOri}
 
Tables \ref{tab:eight-plant} and \ref{tab:eight-drone} report the eight-orientation results. Increasing the number of modeled orientations leads to further accuracy improvements, especially for the plant dataset. Because convolutional pooling over eight orientations produces twice the feature-map dimensionality of the four-orientation configuration. However, it imposes substantially greater memory and computational demands on conventional implementations. The scatter based convolution continues to provide substantial efficiency benefits.
On the plant dataset, scatter is approximately 45.4\% faster than cuDNN at 256 resolution and about 34.5\% faster at 1024 resolution, while reducing energy usage by approximately 41.2\% and 21.7\%, respectively.
On the drone dataset, scatter achieves around 56.9\% faster training at 256 resolution and 32.7\% faster at 1024 resolution, with energy savings of about 44.3\% and 22.7\%, respectively.

The proposed scatter-based approach differs from E2CNN in how rotation is modeled. While E2CNN learns orientation-dependent expansion coefficients, the scatter-based method constructs steerable filters from a set of learnable base filters, as described in Section~\ref{sec:fft_arbitrary}, where orientation dependence is introduced analytically. By enforcing orthogonality and equal-magnitude constraints on these base filters, the formulation approximates rotation invariance, leading to accuracy trends distinct from E2CNN.

The experimental results demonstrate that as rotation resolution increases, rotation invariant convolutions become more accurate, and the scatter based implementation remains more efficient than ORN and E2CNN. ORN shows smaller accuracy improvements in the eight-orientation setting because it does not provide an API for orientation-subgroup pooling; in our implementation, its group pooling operation aggregates all eight orientations into a single pooled feature.
As accuracy continues to improve, conventional rotation-invariant convolutions incur rapidly increasing memory and computation overhead due to expanded feature-map dimensionality.
This effect is particularly evident for E2CNN, where group representations lead to larger intermediate tensors.
By contrast, the proposed scatter-based implementation mitigates this issue by consolidating rotation handling within a single convolution pass, thereby maintaining scalability as orientation resolution increases.

\begin{table}[ht]
\centering
\caption{Plant Data Segmentation Using Various Eight-Orientation Rotation-Invariant Convolutions}
\renewcommand{\arraystretch}{1.2}
\begin{tabular}{c|c|c|c|c} 
\hline\hline
\shortstack{Image\\size} & 
\shortstack{Convolution\\applied} & 
\shortstack{Training\\time (s)} &  
\shortstack{Energy\\(kWh)} &  
\shortstack{Test\\accuracy (\%)} \\ 
\hline
\hline
\multirow{4}{*}{256} 
 & cudnn based conv    & 36866.0  & 1.675  & 78.69 \\ \cline{2-5}
 & ORN conv            & 36188.0  & 1.612  & 76.13 \\ \cline{2-5}
 & E2CNN conv          & 33408.0  & 1.561  &78.15  \\ \cline{2-5}
 & scatter based conv  & 20133.0  & 0.9844  &78.64  \\
\hline 
\multirow{4}{*}{1024} 
  & cudnn based conv    &100269.79  &6.8162  &75.86 \\ \cline{2-5}
  & ORN conv            &95240.30   &6.5967  &74.19 \\ \cline{2-5}
  & E2CNN conv          &84227.06   &5.8908  &75.57 \\ \cline{2-5}
  & scatter based conv  &65703.51   &5.3363  &75.82 \\
\hline

\end{tabular}
\label{tab:eight-plant}
\end{table}

\begin{table}[ht]
\centering
\caption{Semantic drone segmentation Using Various Eight-Orientation Rotation-Invariant Convolutions}
\renewcommand{\arraystretch}{1.2}
\begin{tabular}{c|c|c|c|c} 
\hline\hline
\shortstack{Image\\size} & 
\shortstack{Convolution\\applied} & 
\shortstack{Training\\time (s)} &  
\shortstack{Energy\\(kWh)} &  
\shortstack{Test\\accuracy (\%)} \\ 
\hline
\hline
\multirow{4}{*}{256} 
 & cudnn based conv    &14002.8  &0.6345 &93.98  \\ \cline{2-5}
 & ORN conv    &13798.3  &0.6132  &92.76  \\ \cline{2-5}
 & E2CNN conv    &12682.3 &0.5894  &93.67  \\ \cline{2-5}
 & scatter based conv  &6042.28  & 0.3533 &93.91  \\
\hline
\multirow{4}{*}{1024}
  & cudnn based conv    &41836.42  &2.8440 &92.44 \\ \cline{2-5}
  & ORN conv            &41382.05  &2.8402 &91.20 \\ \cline{2-5}
  & E2CNN conv          &36342.18  &2.4986 &91.99 \\ \cline{2-5}
  & scatter based conv  &28137.53  &2.1996 &92.40 \\
\hline
\end{tabular}
\label{tab:eight-drone}
\end{table}

\subsubsection{Symetric sixteen orientation rotation convolution}\label{subsubsec:sixteenOri}
In preliminary experiments, we found that modeling sixteen orientations in UNet leads to very large intermediate feature tensors and impractical memory and compute overhead. Compared to implementations that process each rotation separately, our scatter-based approach performs symmetric four-rotation convolution and pooling in a single pass within GPU on-chip memory, reducing the number of intermediate feature maps by a factor of four. As a result, our scatter-based implementation is the only approach capable of supporting a full sixteen-orientation convolution and pooling within practical memory and runtime constraint. As shown in Tables \ref{tab:sixteen-plant} and \ref{tab:sixteen-drone}, the scatter-based approach exhibits improved accuracy when employing sixteen orientations relative to the eight-orientation configuration. Because convolutional pooling over sixteen orientations generates twice the feature-map dimensionality of the eight-orientation configuration.

\begin{table}[ht]
\centering
\caption{Plant Data Segmentation Using Sixtee-Orientation Rotation-Invariant Convolution}
\renewcommand{\arraystretch}{1.2}
\begin{tabular}{c|c|c|c|c} 
\hline\hline
\shortstack{Image\\size} & 
\shortstack{Convolution\\applied} & 
\shortstack{Training\\time (s)} &  
\shortstack{Energy\\(kWh)} &  
\shortstack{Test\\accuracy (\%)} \\ 
\hline
\hline
256
 & scatter based conv  & 31208.3  &1.555   &79.27  \\
\hline
1024
 & scatter based conv  &103612.2   &6.685    &76.33  \\
\hline
\end{tabular}
\label{tab:sixteen-plant}
\end{table}

\begin{table}[ht]
\centering
\caption{Semantic drone segmentation Using Sixteen-Orientation Rotation-Invariant Convolution}
\renewcommand{\arraystretch}{1.2}
\begin{tabular}{c|c|c|c|c} 
\hline\hline
\shortstack{Image\\size} & 
\shortstack{Convolution\\applied} & 
\shortstack{Training\\time (s)} &  
\shortstack{Energy\\(kWh)} &  
\shortstack{Test\\accuracy (\%)} \\ 
\hline\hline
256
 & scatter based conv  & 9982.3 &0.513  &94.39  \\
\hline
1024
 & scatter based conv  & 41905.6 &3.101 &92.87  \\
\hline
\end{tabular}
\label{tab:sixteen-drone}
\end{table}

\subsection{Post-Segmentation Processing and Visualization}

\begin{figure}[htbp]                                
  \centering
\begin{subfigure}[t]{0.45\linewidth}
  \includegraphics[width=\linewidth]{\detokenize{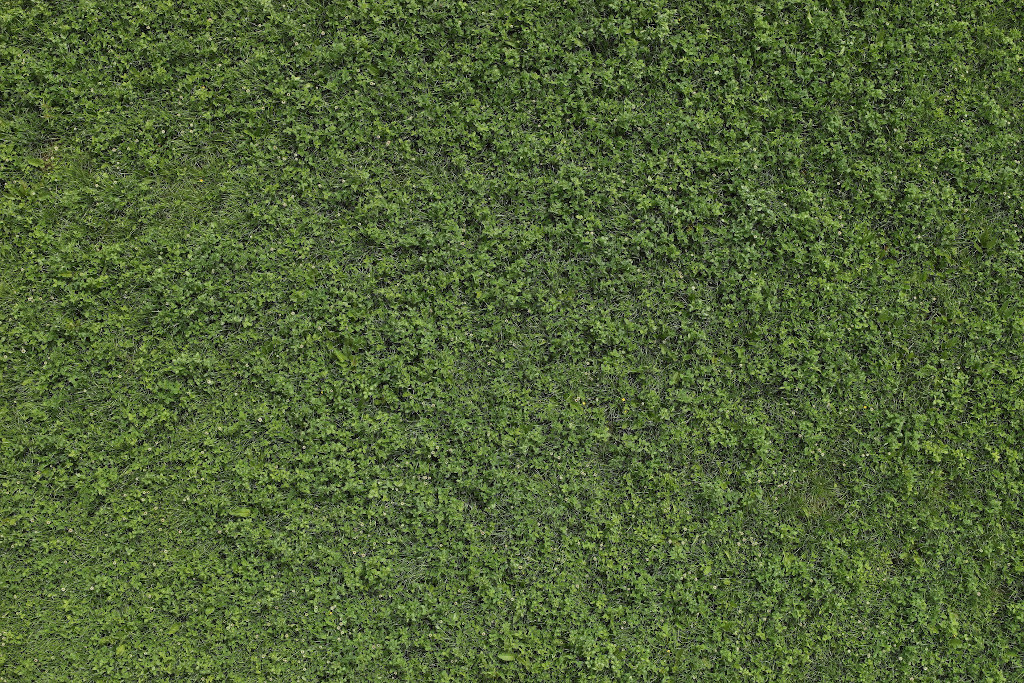}}
  \label{fig:a}
\end{subfigure}
\hspace{0.1em} 
\begin{subfigure}[t]{0.45\linewidth}
  \includegraphics[width=\linewidth]{\detokenize{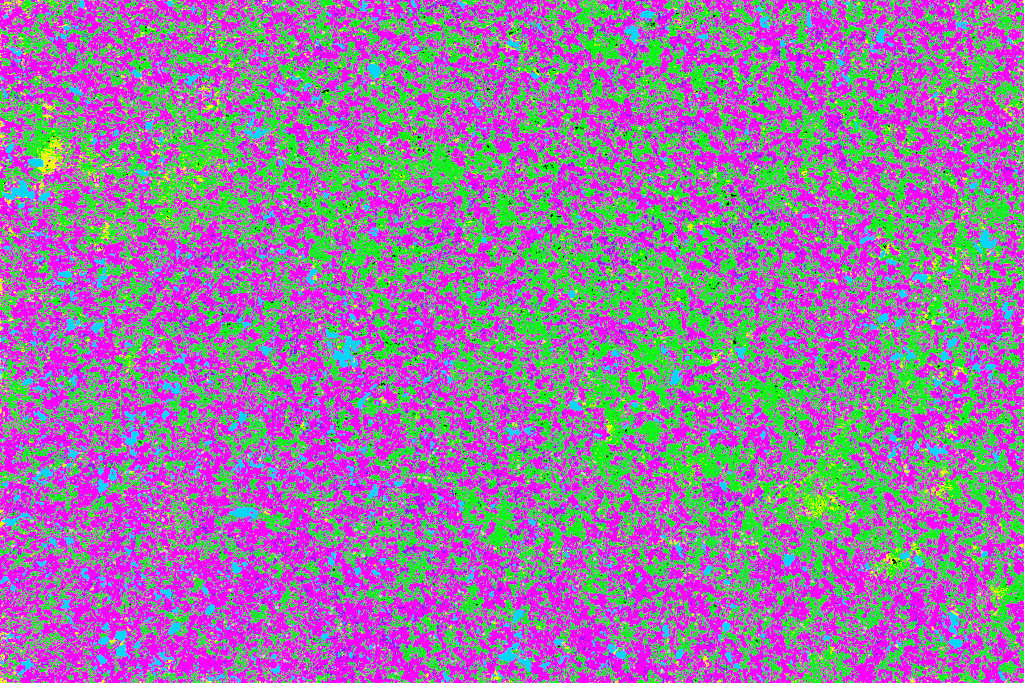}}
  \label{fig:b}
\end{subfigure}
\par\vspace{0.01em}

\begin{subfigure}[t]{0.45\linewidth}
  \includegraphics[width=\linewidth]{\detokenize{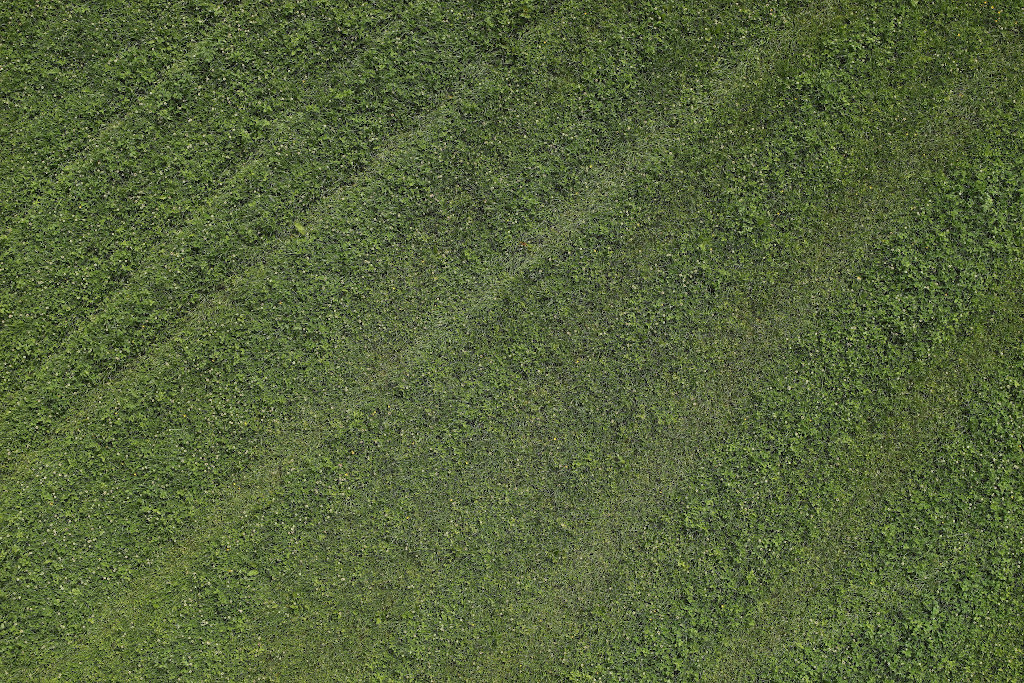}}
  \label{fig:c}
\end{subfigure}
\hspace{0.1em} 
\begin{subfigure}[t]{0.45\linewidth}
  \includegraphics[width=\linewidth]{\detokenize{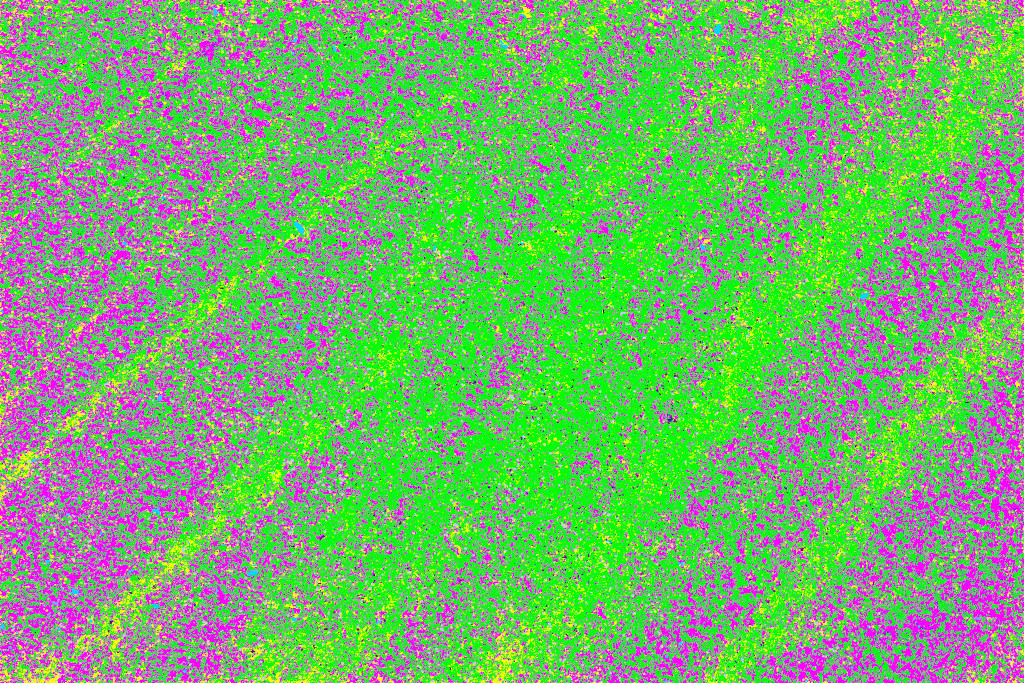}}
  \label{fig:d}
\end{subfigure}
\par\vspace{0.01em}

\begin{subfigure}[t]{0.45\linewidth}
  \includegraphics[width=\linewidth]{\detokenize{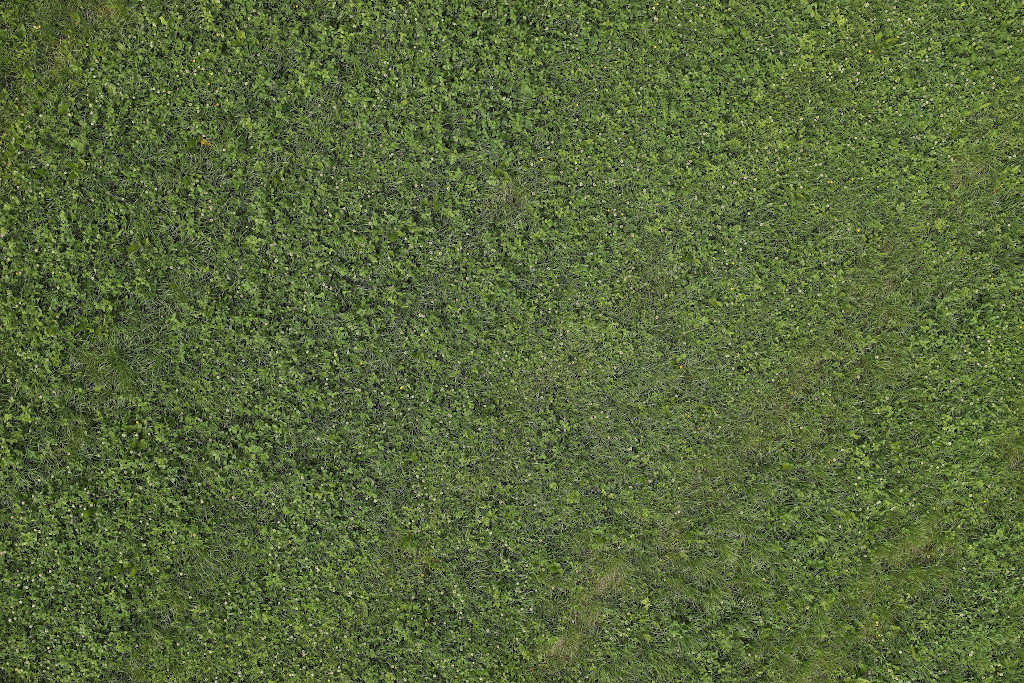}}
  \label{fig:c}
\end{subfigure}
\hspace{0.1em} 
\begin{subfigure}[t]{0.45\linewidth}
  \includegraphics[width=\linewidth]{\detokenize{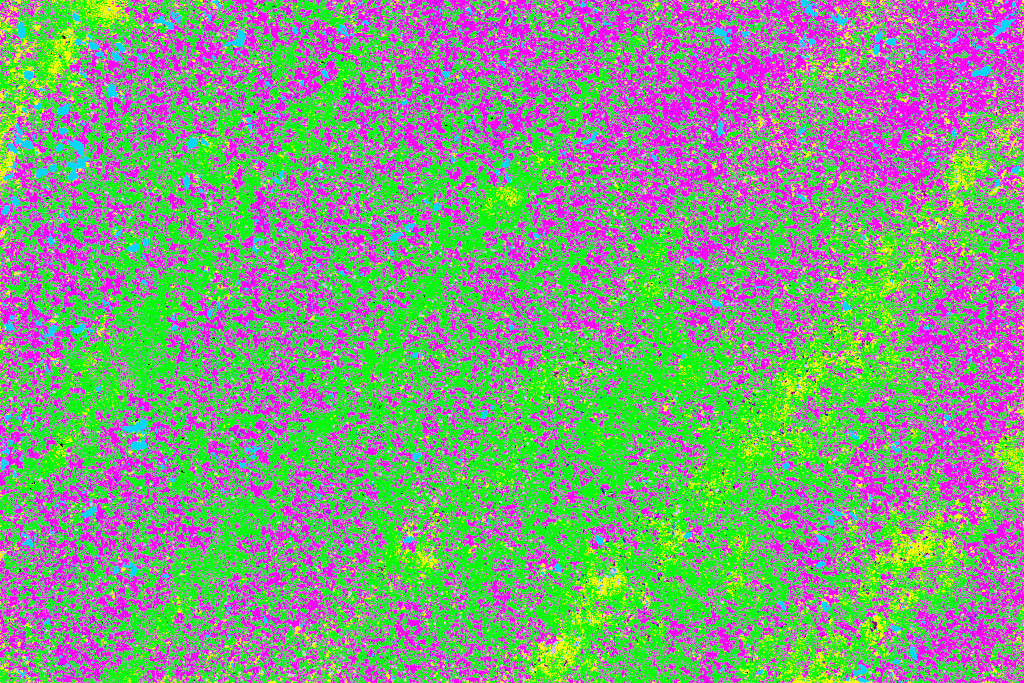}}
  \label{fig:d}
\end{subfigure}
\par\vspace{0.01em}

  \caption{Original P1 RGB imagery (left) and corresponding segmentation masks (right).}
  \label{fig:six-grid}
\end{figure}

\begin{figure}[tbp]                 
  \centering
  \begin{subfigure}[t]{0.90\linewidth}
    \centering
    \includegraphics[width=\linewidth]{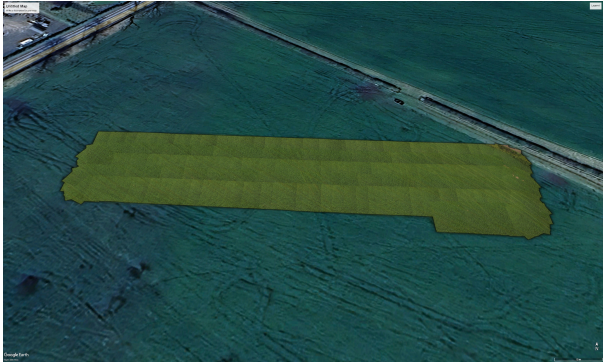}  
    \label{fig:top}
  \end{subfigure}
  \vspace{0.01em}             
  \begin{subfigure}[t]{0.90\linewidth}
    \centering
    \includegraphics[width=\linewidth]{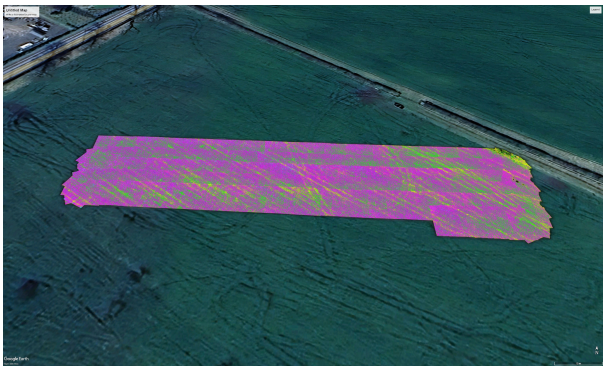} 
    \label{fig:bottom}
  \end{subfigure}

  \caption{Geo-referenced P1 RGB imagery mosaic (Top) and corresponding Geo-referenced mask mosaic (Bottom) displayed in Google Earth Pro.}
  \label{fig:mosaic}
\end{figure}

\begin{figure}[htbp]
  \centering
  \begin{subfigure}{0.44\textwidth}
    \centering
    \includegraphics[width=\textwidth]{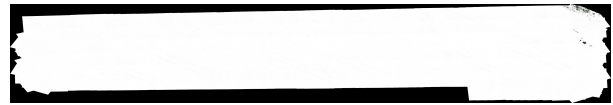}
    \caption{Visualization of DSM (white) and Soil segmentation mask (black inside white DSM). }
    \label{fig:a}
  \end{subfigure}\par\medskip
  \begin{subfigure}{0.44\textwidth}
    \centering
    \includegraphics[width=\textwidth]{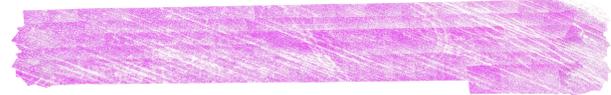}
   	 \caption{Visualization of Clover leaf segmentation mask.}
    \label{fig:b}
  \end{subfigure}\par\medskip
  
  \begin{subfigure}{0.44\textwidth}
    \centering
    \includegraphics[width=\textwidth]{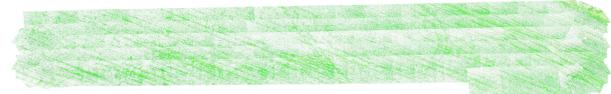}
   	 \caption{Visualization of Grass segmentation mask.}
    \label{fig:c}
  \end{subfigure}\par\medskip

  \begin{subfigure}{0.44\textwidth}
    \centering
    \includegraphics[width=\textwidth]{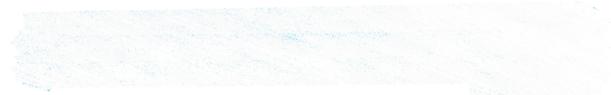}
   	 \caption{Visualization of Dock leaf segmentation mask}
    \label{fig:d}
  \end{subfigure}\par\medskip

  \begin{subfigure}{0.44\textwidth}
    \centering
    \includegraphics[width=\textwidth]{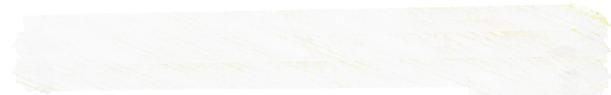}
  	 \caption{Visualization of weeds and other plant segmentation mask.}
    \label{fig:e}
  \end{subfigure}

  \caption{Visualization of individual species masks}
  \label{fig:sep}
\end{figure}

The rotation-invariant U-Net segmentation network was trained to segment plant species in the imagery, 
and the plant-species segmentation masks from all frames (see Fig.~\ref{fig:six-grid}) were subsequently georeferenced and mosaicked into a seamless orthomosaic.
The resulting species‑labeled mosaic was exported to Google Earth Pro for visualization (Fig.~\ref{fig:mosaic}) and for analysis of species distribution across the field. 
As shown in Fig.~\ref{fig:sep}, the plant area consists of 51.41\% clover leaf, 37.89\% grass, 3.13\% dock leaf, and 7.56\% weeds and other vegetation, 
indicating clover as the dominant ground cover followed by grass, with minimal presence of dock leaves and miscellaneous weeds.

\section{Conclusion}
In conclusion, replacing cuDNN’s dense kernels with our single-pass scatter convolution accelerates both standard (single-orientation) and rotation-invariant layers without affecting accuracy. For standard convolutions, the scatter kernel lowers memory pressure and computational load, reducing training time and energy consumption on small- and medium-sized feature maps with deep channels.
When rotations are required it fuses all filters into one pass, yielding even larger speed-and-energy gains. Results on the Semantic Drone and plant-segmentation benchmarks confirm consistent improvements across input sizes, establishing scatter convolution as a practical drop-in upgrade for modern CNNs.

\section*{Acknowledgments}
\textcolor{black}{This work is funded by the UK Natural Environment Research Council under project FLORA-SAGE - Federated Learning Optimised for Remote Assessment of Species in Agricultural Grassland Ecosystems (grant award number UKRI053) and UKRI Digital Dairy Value-Chain for South-West Scotland and Cumbria (Strength in Places Fund) (grant award number 99890)}

\appendices
\section{Performance of Scatter-Based Convolution Compared to cuDNN Convolutions} \label{appendix:compareCUDNN}

In the single-rotation case, the rotation-invariant convolution reduces to a standard convolution. We evaluate our scatter-based convolution (single orientation) through a series of experiments and benchmark it against NVIDIA’s cuDNN convolution library, which is widely used in deep-learning applications \cite{cuDNNsdk}. In our implementation, the fastest cuDNN kernel is automatically selected by the cuDNN autotuner.
The GPU used for evaluation is NVIDIA GeForce RTX 3080 Ti with 16GB video memory. The version of CUDA 12.3 and cuDNN 8.9.6.50 are installed in a Windows 10 system. 

\vspace{0.4cm}  
\centerline{
\begin{tikzpicture}
\begin{axis}[
width=9cm,
height=5cm,
xlabel={\shortstack{Primary X-axis: Number of filters \\ Secondary X-axis: Number of input channels}},
xlabel style={font=\scriptsize},
ylabel={Run time (ms)},
ylabel style={font=\scriptsize},
xtick={1,6,11,16,21,26,31,36,41,46,51,56,61,66,71,76,81,86,91,96,101},
xticklabels={4,8,16,32,64,128,{256\\~256},{4},{8},{16},{32},{64},{128},{512\\~256},{4},{8},{16},{32},{64},{128},{1024\\~256}},
x tick label style={font=\tiny, rotate=90},
y tick label style={font=\tiny},
ymin=0,
ymax=0.35,
legend pos=north west,
legend style={font=\tiny},
xtick pos=bottom,
ytick pos=left,
grid=major,
grid style={dashed,gray!30},
]

\addplot[
color=blue,
mark=star,
mark size=2pt,
line width=1.0pt,
mark options={solid, fill=blue!70}
] coordinates {
(1,0.025) (6,0.028) (11,0.035) (16,0.052) (21,0.055) (26,0.058) (31,0.125)
(36,0.052) (41,0.055) (46,0.058) (51,0.115) (56,0.128) (61,0.152) (66,0.210)
(71,0.105) (76,0.108) (81,0.115) (86,0.188) (91,0.195) (96,0.208) (101,0.305)
};

\addplot[
color=orange,
mark=triangle,
mark size=2pt,
dashed,
line width=1.0pt,
mark options={solid, fill=orange!70}
] coordinates {
(1,0.022) (6,0.025) (11,0.032) (16,0.040) (21,0.052) (26,0.080) (31,0.120)
(36,0.025) (41,0.028) (46,0.032) (51,0.045) (56,0.068) (61,0.115) (66,0.202)
(71,0.042) (76,0.045) (81,0.048) (86,0.065) (91,0.095) (96,0.165) (101,0.288)
};

\legend{CUDNN Fastest Conv, Conv With Scatter Operation}

\end{axis}
\node[below=50pt] at (current axis.south) {\normalsize Fig.\ref{fig:performance} (a) $4\times4$ input};
\end{tikzpicture}
}
\vspace{0.4cm}  

\centerline{
\begin{tikzpicture}
\begin{axis}[
width=9cm,
height=5cm,
xlabel={\shortstack{Primary X-axis: Number of filters \\ Secondary X-axis: Number of input channels}},
xlabel style={font=\scriptsize},
ylabel={Run time (ms)},
ylabel style={font=\scriptsize},
xtick={1,6,11,16,21,26,31,36,41,46,51,56,61,66,71,76,81,86,91,96,101},
xticklabels={4,8,16,32,64,128,{256\\~256},{4},{8},{16},{32},{64},{128},{512\\~256},{4},{8},{16},{32},{64},{128},{1024\\~256}},
x tick label style={font=\tiny, rotate=90},
y tick label style={font=\tiny},
ymin=0,
ymax=0.9,
legend pos=north west,
legend style={font=\tiny},
xtick pos=bottom,
ytick pos=left,
grid=major,
grid style={dashed,gray!30},
]

\addplot[
color=blue,
mark=star,
mark size=2pt,
line width=1.0pt,
mark options={solid, fill=blue!70}
] coordinates {
(1,0.075) (6,0.085) (11,0.090) (16,0.105) (21,0.135) (26,0.240) (31,0.145)
(36,0.145) (41,0.160) (46,0.185) (51,0.250) (56,0.275) (61,0.390) (66,0.230)
(71,0.235) (76,0.255) (81,0.365) (86,0.445) (91,0.515) (96,0.775) (101,0.785)
};

\addplot[
color=orange,
mark=triangle,
mark size=2pt,
dashed,
line width=1.0pt,
mark options={solid, fill=orange!70}
] coordinates {
(1,0.045) (6,0.055) (11,0.065) (16,0.085) (21,0.130) (26,0.185) (31,0.065)
(36,0.090) (41,0.085) (46,0.120) (51,0.180) (56,0.240) (61,0.390) (66,0.095)
(71,0.125) (76,0.135) (81,0.210) (86,0.255) (91,0.485) (96,0.725) (101,0.730)
};

\legend{CUDNN Fastest Conv, Conv With Scatter Operation}

\end{axis}
\node[below=50pt] at (current axis.south) {\normalsize Fig.\ref{fig:performance} (b) $8\times8$ input};
\end{tikzpicture}
}
\vspace{0.4cm}  

\centerline{
\begin{tikzpicture}
\begin{axis}[
width=9cm,
height=5cm,
xlabel={\shortstack{Primary X-axis: Number of filters \\ Secondary X-axis: Number of input channels}},
xlabel style={font=\scriptsize},
ylabel={Run time (ms)},
ylabel style={font=\scriptsize},
xtick={1,6,11,16,21,26,31,36,41,46,51,56,61,66,71,76,81,86,91,96,101},
xticklabels={4,8,16,32,64,128,{256\\~256},{4},{8},{16},{32},{64},{128},{512\\~256},{4},{8},{16},{32},{64},{128},{1024\\~256}},
x tick label style={font=\tiny, rotate=90},
y tick label style={font=\tiny},
ymin=0,
ymax=2.5,
legend pos=north west,
legend style={font=\tiny},
xtick pos=bottom,
ytick pos=left,
grid=major,
grid style={dashed,gray!30},
]

\addplot[
color=blue,
mark=star,
mark size=2pt,
line width=1.0pt,
mark options={solid, fill=blue!70}
] coordinates {
(1,0.15) (6,0.15) (11,0.15) (16,0.15) (21,0.25) (26,0.45) (31,0.72)
(36,0.28) (41,0.32) (46,0.28) (51,0.38) (56,0.52) (61,0.77) (66,1.35)
(71,0.58) (76,0.58) (81,0.58) (86,0.60) (91,1.05) (96,1.28) (101,2.02)
};

\addplot[
color=orange,
mark=triangle,
mark size=2pt,
dashed,
line width=1.0pt,
mark options={solid, fill=orange!70}
] coordinates {
(1,0.08) (6,0.12) (11,0.15) (16,0.22) (21,0.35) (26,0.48) (31,0.95)
(36,0.15) (41,0.20) (46,0.25) (51,0.40) (56,0.52) (61,0.70) (66,1.28)
(71,0.38) (76,0.42) (81,0.50) (86,0.65) (91,0.95) (96,1.28) (101,2.00)
};

\legend{CUDNN Fastest Conv, Conv With Scatter Operation}

\end{axis}
\node[below=50pt] at (current axis.south) {\normalsize Fig.\ref{fig:performance} (c) $16\times16$ input};
\end{tikzpicture}
}
\vspace{0.4cm}  

\centerline{
\begin{tikzpicture}
\begin{axis}[
width=9cm,
height=5cm,
xlabel={\shortstack{Primary X-axis: Number of filters \\ Secondary X-axis: Number of input channels}},
xlabel style={font=\scriptsize},
ylabel={Run time (ms)},
ylabel style={font=\scriptsize},
xtick={1,6,11,16,21,26,31,36,41,46,51,56,61,66,71,76,81,86,91,96,101},
xticklabels={4,8,16,32,64,128,{256\\~256},{4},{8},{16},{32},{64},{128},{512\\~256},{4},{8},{16},{32},{64},{128},{1024\\~256}},
x tick label style={font=\tiny, rotate=90},
y tick label style={font=\tiny},
ymin=0,
ymax=9,
legend pos=north west,
legend style={font=\tiny},
xtick pos=bottom,
ytick pos=left,
grid=major,
grid style={dashed,gray!30},
]

\addplot[
color=blue,
mark=star,
mark size=2pt,
line width=1.0pt,
mark options={solid, fill=blue!70}
] coordinates {
(1,0.6) (6,0.6) (11,0.6) (16,0.6) (21,1.1) (26,1.7) (31,2.7)
(36,1.1) (41,1.1) (46,1.1) (51,1.2) (56,1.9) (61,2.8) (66,4.2)
(71,2.0) (76,1.9) (81,2.1) (86,2.2) (91,4.3) (96,5.3) (101,8.2)
};

\addplot[
color=orange,
mark=triangle,
mark size=2pt,
dashed,
line width=1.0pt,
mark options={solid, fill=orange!70}
] coordinates {
(1,0.3) (6,0.4) (11,0.5) (16,0.7) (21,1.0) (26,1.8) (31,3.3)
(36,0.7) (41,0.8) (46,1.0) (51,1.2) (56,1.9) (61,2.7) (66,4.8)
(71,1.4) (76,1.5) (81,1.7) (86,2.1) (91,3.1) (96,4.5) (101,8.0)
};

\legend{CUDNN Fastest Conv, Conv With Scatter Operation}

\end{axis}
\node[below=50pt] at (current axis.south) {\normalsize Fig.\ref{fig:performance} (d) $32\times32$ input};
\end{tikzpicture}
}
\vspace{0.4cm}  

\centerline{
\begin{tikzpicture}
\begin{axis}[
width=9cm,
height=5cm,
xlabel={\shortstack{Primary X-axis: Number of filters \\ Secondary X-axis: Number of input channels}},
xlabel style={font=\scriptsize},
ylabel={Run time (ms)},
ylabel style={font=\scriptsize},
xtick={1,6,11,16,21,26,31,36,41,46,51,56,61,66,71,76,81,86,91,96,101},
xticklabels={4,8,16,32,64,128,{256\\~256},{4},{8},{16},{32},{64},{128},{512\\~256},{4},{8},{16},{32},{64},{128},{1024\\~256}},
x tick label style={font=\tiny, rotate=90},
y tick label style={font=\tiny},
ymin=0,
ymax=40,
legend pos=north west,
legend style={font=\tiny},
xtick pos=bottom,
ytick pos=left,
grid=major,
grid style={dashed,gray!30},
]

\addplot[
color=blue,
mark=star,
mark size=2pt,
line width=1.0pt,
mark options={solid, fill=blue!70}
] coordinates {
(1,2.5) (6,2.8) (11,2.8) (16,3.2) (21,4.2) (26,6.2) (31,10.2)
(36,5.0) (41,5.2) (46,4.8) (51,5.0) (56,7.2) (61,11.2) (66,18.0)
(71,9.8) (76,9.8) (81,10.5) (86,10.2) (91,18.8) (96,21.2) (101,35.0)
};

\addplot[
color=orange,
mark=triangle,
mark size=2pt,
dashed,
line width=1.0pt,
mark options={solid, fill=orange!70}
] coordinates {
(1,1.5) (6,1.8) (11,2.2) (16,3.0) (21,4.5) (26,7.8) (31,12.2)
(36,2.8) (41,3.2) (46,3.5) (51,4.8) (56,7.0) (61,12.0) (66,19.8)
(71,5.0) (76,5.2) (81,6.2) (86,8.2) (91,12.8) (96,19.8) (101,33.8)
};

\legend{CUDNN Fastest Conv, Conv With Scatter Operation}

\end{axis}
\node[below=50pt] at (current axis.south) {\normalsize Fig.\ref{fig:performance} (e) $64\times64$ input};
\end{tikzpicture}
}
\vspace{0.4cm}  

\centerline{
\begin{tikzpicture}
\begin{axis}[
width=9cm,
height=5cm,
xlabel={\shortstack{Primary X-axis: Number of filters \\ Secondary X-axis: Number of input channels}},
xlabel style={font=\scriptsize},
ylabel={Run time (ms)},
ylabel style={font=\scriptsize},
xtick={1,2,3,4,5,7,8,9,10,11,13,14,15,16,17},
xticklabels={4,8,16,32,{64\\~64},{4},{8},{16},{32},{128\\~64},{4},{8},{16},{32},{256\\~64}},
x tick label style={font=\tiny, rotate=90},
y tick label style={font=\tiny},
ymin=0,
ymax=4.5,
legend pos=north west,
legend style={font=\tiny},
xtick pos=bottom,
ytick pos=left,
grid=major,
grid style={dashed,gray!30},
]

\addplot[
color=blue,
mark=star,
mark size=2pt,
line width=1.0pt,
mark options={solid, fill=blue!70}
] coordinates {
(1,0.75) (2,0.75) (3,0.7) (4,0.75) (5,1.25)
(7,1.3) (8,1.3) (9,1.35) (10,1.4) (11,1.95)
(13,2.4) (14,2.4) (15,2.45) (16,2.7) (17,4.25)
};

\addplot[
color=orange,
mark=triangle,
mark size=2pt,
dashed,
line width=1.0pt,
mark options={solid, fill=orange!70}
] coordinates {
(1,0.65) (2,0.75) (3,1.05) (4,1.5) (5,2.4)
(7,0.9) (8,1.05) (9,1.3) (10,1.9) (11,3.2)
(13,1.5) (14,1.65) (15,2.05) (16,2.5) (17,4.1)
};

\legend{CUDNN Fastest Conv, Conv With Scatter Operation}

\end{axis}
\node[below=45pt] at (current axis.south) {\normalsize Fig.\ref{fig:performance} (f) $128\times128$ input};
\end{tikzpicture}
}

\begin{figure}[htbp]
\centering
\begin{tikzpicture}
\begin{axis}[
width=9cm,
height=5cm,
xlabel={\shortstack{Primary X-axis: Number of filters \\ Secondary X-axis: Number of input channels}},
xlabel style={font=\scriptsize},
ylabel={Run time (ms)},
ylabel style={font=\scriptsize},
xtick={1,2,3,4,5,7,8,9,10,11,13,14,15,16,17},
xticklabels={4, 8, 16, 32, \rotatebox{90}{\rotatebox{-90}{\shortstack{64  64}}}, 4, 8, 16, 32, \rotatebox{90}{\rotatebox{-90}{\shortstack{128  64}}}, 4, 8, 16, 32, \rotatebox{90}{\rotatebox{-90}{\shortstack{256  64}}}},
x tick label style={font=\tiny, rotate=90},
y tick label style={font=\tiny},
ymin=0,
ymax=20,
legend pos=north west,
legend style={font=\tiny},
xtick pos=bottom,
ytick pos=left,
grid=major,
grid style={dashed,gray!30},
]

\addplot[
color=blue,
mark=star,
mark size=2pt,
line width=1.0pt,
mark options={solid, fill=blue!70}
] coordinates {
(1,2.5) (2,2.5) (3,2.8) (4,2.8) (5,4.0)
(7,5.5) (8,5.2) (9,5.2) (10,5.2) (11,9.0)
(13,9.7) (14,9.8) (15,9.8) (16,11.3) (17,17.7)
};

\addplot[
color=orange,
mark=triangle,
mark size=2pt,
dashed,
line width=1.0pt,
mark options={solid, fill=orange!70}
] coordinates {
(1,2.5) (2,2.5) (3,4.0) (4,6.2) (5,9.5)
(7,4.0) (8,4.8) (9,5.2) (10,7.0) (11,12.5)
(13,7.5) (14,7.5) (15,8.5) (16,11.0) (17,16.8)
};

\legend{CUDNN Fastest Conv, Conv With Scatter Operation}

\end{axis}
\node[below=45pt] at (current axis.south) {\normalsize Fig.\ref{fig:performance}  (g) $256\times256$ input};
\end{tikzpicture}
\caption{Performance comparison of cuDNN (blue) and
the proposed scatter convolution (orange) across seven
input resolutions.}
\label{fig:performance}
\end{figure}

NVIDIA's cuDNN offers eight distinct forward-convolution
implementations, and for every experiment we measure the fastest of
those eight via cuDNN's autotuner. As shown in Fig.\ref{fig:performance} (a)-(g),
across the seven plots (input sizes $4 \to 256$) our
scatter-based kernel still dominates that ``best-of-eight'' cuDNN
baseline in the majority of layer configurations. Counting every
(resolution $\times$ input-channel $\times$ filter) triplet, the scatter
variant is quicker in most cases, with typical speed-ups of
$1.1\times$--$1.6\times$ and a peak of $\approx 2\times$ for the very
small $4\times4$ activations. The gain is most pronounced whenever the
spatial footprint is modest ($\le 32\times32$) or the channel--filter
product is under $\sim16\text{\,K}$, where kernel-launch latency and
global-memory traffic dominate the cost of convolution; the single-pass
scatter design removes the data-lowering staging step and sustains
higher memory-bandwidth utilisation in that regime. As the resolution
grows beyond $64\times64$ and the workload becomes compute-bound,
cuDNN paths occasionally pull ahead, yet even at
$256\times256$ our kernel still wins for 4 of the 9 channel--filter
pairs tested. In practice, this means the scatter kernel is the better
choice for mid‑to‑deep layers where spatial dimensions have shrunk and channel depth has increased, while cuDNN
remains preferable only for the first one or two encoder layers of very
high-resolution networks.

\bibliographystyle{IEEEtran} 
\bibliography{relatedworks.bib} 

@article{FFT1,
author={T. {Abtahi} and C. {Shea} and A. {Kulkarni} and T. {Mohsenin}}, 
journal={IEEE Transactions on Very Large Scale Integration (VLSI) Systems}, 
title={{Accelerating Convolutional Neural Network With FFT on Embedded Hardware}}, 
year={2018}, 
volume={26}, 
number={9}, 
pages={1737-1749}, 
month={9},
}

@article{FFT2,
  title={{Fast convolutional nets with fbfft: A GPU performance evaluation}},
  author={Vasilache Nicolas and Johnson Jeff and Mathieu Michael and Chintala Soumith and Piantino Serkan and LeCun Yann},
  journal={arXiv:1412.7580},
  year={2014},
}

@inproceedings{FFT3,
title = {{ZNN - A Fast and Scalable Algorithm for Training 3D Convolutional Networks on Multi-core and Many-Core Shared Memory Machines}},
author = {Aleksandar Zlateski and Kisuk Lee and H. Sebastian Seung},
year = {2016},
day = {18},
booktitle = {2016 IEEE 30th International Parallel and Distributed Processing Symposium(IPDPS)},
pages = {801--811},
}

@inproceedings{FFT4,
title = {{Fast training of convolutional networks through FFTs}},
author = {Micha{\"e}l Mathieu and Mikael Henaff and Yann Lecun},
year = {2014},
booktitle = {International Conference on Learning Representations (ICLR2014)},
}

@article{cuDNN,
  author    = {Sharan Chetlur and
               Cliff Woolley and
               Philippe Vandermersch and
               Jonathan Cohen and
               John Tran and
               Bryan Catanzaro and
               Evan Shelhamer},
  title     = {{cuDNN: Efficient Primitives for Deep Learning}},
  journal   = {CoRR},
  volume    = {abs/1410.0759},
  year      = {2014},
}

@article{Lavin,
  title={{Fast Algorithms for Convolutional Neural Networks}},
  author={Andrew Lavin and Scott Gray},
  journal={2016 IEEE Conference on Computer Vision and Pattern Recognition (CVPR)},
  year={2015},
  pages={4013-4021},
}

@Book{winograd,
author = { Winograd, S. },
title = { Arithmetic complexity of computations / Shmuel Winograd },
isbn = { 0898711630},
publisher = {Society for Industrial and Applied Mathematics Philadelphia},
year = { 1980 },
type = { Book },
language = { English },
}

@InProceedings{G,
  title = 	 {{Group Equivariant Convolutional Networks}},
  author = 	 {Taco Cohen and Max Welling},
  booktitle = 	 {Proceedings of The 33rd International Conference on Machine Learning},
  pages = 	 {2990--2999},
  year = 	       {2016},
  volume = 	 {48},
}

@article{featureTrans,
  title={Exploiting cyclic symmetry in convolutional neural networks},
  author={Dieleman, Sander and De Fauw, Jeffrey and Kavukcuoglu, Koray},
  journal={arXiv:1602.02660},
  year={2016}
}

@inproceedings{hexa,
title	= {{HexaConv}},
author	= {Emiel Hoogeboom and Jorn W T Peters and Taco S Cohen and Max Welling},
year	= {2018},
booktitle	= {International Conference on Machine Learning (ICML)}
}

@article{cohen2018general,
  title={{A general theory of equivariant cnns on homogeneous spaces}},
  author={Cohen, Taco and Geiger, Mario and Weiler, Maurice},
  journal={arXiv:1811.02017},
  year={2018}
}

@article{orn,
  title={{Oriented Response Networks}},
  author={Yanzhao Zhou and Qixiang Ye and Qiang Qiu and Jianbin Jiao},
  journal={2017 IEEE Conference on Computer Vision and Pattern Recognition (CVPR)},
  year={2017},
  pages={4961-4970}
}

@article{G-cnn7,
  title={{Learning Steerable Filters for Rotation Equivariant CNNs}},
  author={Maurice Weiler and Fred A. Hamprecht and Martin Storath},
  journal={2018 IEEE/CVF Conference on Computer Vision and Pattern Recognition},
  year={2017},
  pages={849-858}
}

@InProceedings{G-cnn10,
  title = 	 {{Gauge Equivariant Convolutional Networks and the Icosahedral CNN}},
  author = 	 {Cohen, Taco and Weiler, Maurice and Kicanaoglu, Berkay and Welling, Max},
  booktitle = 	 {Proceedings of the 36th International Conference on Machine Learning},
  pages = 	 {1321--1330},
  year = 	 {2019},
  volume = 	 {97},
}

@InProceedings{CubeNet1,
author="Worrall, Daniel
and Brostow, Gabriel",
title={{CubeNet: Equivariance to 3D Rotation and Translation}},
booktitle="ECCV 2018",
pages="585--602",
}

@article{CubeNet2,
  title={{3D G-cnns for pulmonary nodule detection}},
  author={Winkels, Marysia and Cohen, Taco S},
  journal={arXiv:1804.04656},
  year={2018}
}

@inproceedings{equi-theory,
  title={{Equivariance through parameter-sharing}},
  author={Ravanbakhsh, Siamak and Schneider, Jeff and Poczos, Barnabas},
  booktitle={Proceedings of the 34th International Conference on Machine Learning-Volume 70},
  pages={2892--2901},
  year={2017},
}

@inproceedings{shallowNet,
  title={{Learning rotation invariant convolutional filters for texture classification}},
  author={Marcos, Diego and Volpi, Michele and Tuia, Devis},
  booktitle={23rd International Conference on Pattern Recognition (ICPR)},
  pages={2012--2017},
  year={2016},
}

@article{cuDNNsdk,
author = {{NVIDIA Cooperation}},
title = {{cuDNN} Developer Guide},
year = {2019},
url={https://docs.nvidia.com/deeplearning/sdk/cudnn-archived/cudnn\_761/pdf/cuDNN-Developer-Guide.pdf}
}

@inproceedings{harmonic-cnn,
  title={{Harmonic networks: Deep translation and rotation equivariance}},
  author={Worrall, Daniel E and Garbin, Stephan J and Turmukhambetov, Daniyar and Brostow, Gabriel J},
  booktitle={Proceedings of the IEEE Conference on Computer Vision and Pattern Recognition},
  pages={5028--5037},
  year={2017}
}

@article{gabor,
  title={{Gabor convolutional networks}},
  author={Luan, Shangzhen and Chen, Chen and Zhang, Baochang and Han, Jungong and Liu, Jianzhuang},
  journal={IEEE Transactions on Image Processing},
  volume={27},
  number={9},
  pages={4357--4366},
  year={2018},
}

@article{weldon1996efficient,
  title={{Efficient Gabor filter design for texture segmentation}},
  author={Weldon, Thomas P and Higgins, William E and Dunn, Dennis F},
  journal={Pattern recognition},
  volume={29},
  number={12},
  pages={2005--2015},
  year={1996},
}

@article{cheng2018rotdcf,
    title={{RotDCF: Decomposition of Convolutional Filters for Rotation-Equivariant Deep Networks}},
    author={Xiuyuan Cheng and Qiang Qiu and Robert Calderbank and Guillermo Sapiro},
    year={2018},
    journal={arXiv:1805.06846},
}

@article{cudnn-arxiv,
  title={cudnn: Efficient primitives for deep learning},
  author={Chetlur, Sharan and Woolley, Cliff and Vandermersch, Philippe and Cohen, Jonathan and Tran, John and Catanzaro, Bryan and Shelhamer, Evan},
  journal={arXiv preprint arXiv:1410.0759},
  year={2014}
}

@article{cublas,
author = {{NVIDIA Cooperation}},
title = {{cuBLAS} Documentation},
year = {2023},
url={https://docs.nvidia.com/cuda/cublas/}
}

@inproceedings{torchsparse,  
  title={TorchSparse++: Efficient Training and Inference Framework for Sparse Convolution on GPUs},  
  author={Tang, Haotian and Yang, Shang and Liu, Zhijian and Hong, Ke and Yu, Zhongming and Li, Xiuyu and Dai, Guohao and Wang, Yu and Han, Song},  
  booktitle={IEEE/ACM International Symposium on Microarchitecture (MICRO)},  
  year={2023}
}

@article{cnn-survey,
  title={A survey of convolutional neural networks: analysis, applications, and prospects},
  author={Li, Zewen and Liu, Fan and Yang, Wenjie and Peng, Shouheng and Zhou, Jun},
  journal={{IEEE transactions on neural networks and learning systems}},
  volume={33},
  number={12},
  pages={6999--7019},
  year={2021},
  publisher={IEEE}
}

@inproceedings{lu2023im2win,
  title={Im2win: An Efficient Convolution Paradigm on GPU},
  author={Lu, Shuai and Chu, Jun and Guo, Luanzheng and Liu, Xu T},
  booktitle={European Conference on Parallel Processing},
  pages={592--607},
  year={2023},
  organization={Springer}
}

@inproceedings{10.5555/3213069.3213072,
author = {Chang, Qiong and Onishi, Masaki and Maruyama, Tsutomu},
title = {Fast convolution kernels on pascal GPU with high memory efficiency},
year = {2018},
address = {San Diego, CA, USA},
booktitle = {Proceedings of the High Performance Computing Symposium},
articleno = {3},
numpages = {12},
location = {Baltimore, Maryland},
series = {HPC '18}
}

@INPROCEEDINGS{9235051,
  author={Anderson, Andrew and Vasudevan, Aravind and Keane, Cormac and Gregg, David},
  booktitle={2020 IEEE 32nd International Symposium on Computer Architecture and High Performance Computing (SBAC-PAD)}, 
  title={High-Performance Low-Memory Lowering: GEMM-based Algorithms for DNN Convolution}, 
  year={2020},
  volume={},
  number={},
  pages={99-106},
  doi={10.1109/SBAC-PAD49847.2020.00024}}

@ARTICLE{93808,
  author={Freeman, W.T. and Adelson, E.H.},
  journal={IEEE Transactions on Pattern Analysis and Machine Intelligence}, 
  title={The design and use of steerable filters}, 
  year={1991},
  volume={13},
  number={9},
  pages={891-906},
 }

@online{semantic_drone_dataset,
  author       = {Awsaf},
  title        = {Semantic Drone Dataset},
  year         = {2020},
  url          = {https://www.kaggle.com/datasets/awsaf49/semantic-drone-dataset},
  organization = {Kaggle},
}

@online{dji_m300,
  author       = {{DJI}},
  title        = {Matrice 300 RTK},
  year         = {2020},
  organization = {DJI Enterprise},
  url          = {https://www.dji.com/matrice-300},
}

@online{dji_p1,
  author       = {{DJI}},
  title        = {Zenmuse P1 photogrammetry camera},
  year         = {2025},
  organization = {DJI Enterprise},
  url          = {https://enterprise.dji.com/zenmuse-p1},
}

@ARTICLE{10286398,
  author={Fan, Zhihua and Li, Wenming and Wang, Zhen and Liu, Tianyu and Wu, Haibin and Liu, Yanhuan and Wu, Meng and Wu, Xinxin and Ye, Xiaochun and Fan, Dongrui and Sun, Ninghui and An, Xuejun},
  journal={IEEE Transactions on Parallel and Distributed Systems}, 
  title={Accelerating Convolutional Neural Networks by Exploiting the Sparsity of Output Activation}, 
  year={2023},
  volume={34},
  number={12},
  pages={3253-3265}}

@misc{manduhu2025airbornesensedetectdrones,
      title={Airborne Sense and Detect of Drones using Deep Learning and LiDAR Point Clouds}, 
      author={Manduhu Manduhu and Alexander Dow and Petar Trslic and Gerard Dooly and Benjamin Blanck and James Riordan},
      year={2025},
      eprint={2310.09589},
      archivePrefix={arXiv},
      primaryClass={cs.RO},
      url={https://arxiv.org/abs/2310.09589}, 
}

@inproceedings{ronneberger2015u,
  title={U-net: Convolutional networks for biomedical image segmentation},
  author={Ronneberger, Olaf and Fischer, Philipp and Brox, Thomas},
  booktitle={Medical image computing and computer-assisted intervention--MICCAI 2015: 18th international conference, Munich, Germany, October 5-9, 2015, proceedings, part III 18},
  pages={234--241},
  year={2015},
  organization={Springer}
}

@ARTICLE{satellite2,
  author={Zheng, Shangdong and Wu, Zebin and Du, Qian and Xu, Yang and Wei, Zhihui},
  journal={IEEE Transactions on Geoscience and Remote Sensing}, 
  title={Oriented Object Detection for Remote Sensing Images via Object-Wise Rotation-Invariant Semantic Representation}, 
  year={2024},
  volume={62},
  number={},
  pages={1-15},
 }

@ARTICLE{satellite3,
  author={Mei, Shaohui and Jiang, Ruoqiao and Ma, Mingyang and Song, Chao},
  journal={IEEE Transactions on Geoscience and Remote Sensing}, 
  title={Rotation-Invariant Feature Learning via Convolutional Neural Network With Cyclic Polar Coordinates Convolutional Layer}, 
  year={2023},
  volume={61},
  number={},
  pages={1-13},
}

@ARTICLE{satellite4,
  author={Shamsolmoali, Pourya and Zareapoor, Masoumeh and Chanussot, Jocelyn and Zhou, Huiyu and Yang, Jie},
  journal={IEEE Transactions on Geoscience and Remote Sensing}, 
  title={Rotation Equivariant Feature Image Pyramid Network for Object Detection in Optical Remote Sensing Imagery}, 
  year={2022},
  volume={60},
  number={},
  pages={1-14},
  doi={10.1109/TGRS.2021.3112481}}

@ARTICLE{UAV1,
  author={Chen, Yuan and Jiang, Jie},
  journal={IEEE Transactions on Geoscience and Remote Sensing}, 
  title={An Oblique-Robust Absolute Visual Localization Method for GPS-Denied UAV With Satellite Imagery}, 
  year={2024},
  volume={62},
  number={},
  pages={1-13},
 }

@INPROCEEDINGS{geo1,
  author={Tian, Yuxin and Deng, Xueqing and Zhu, Yi and Newsam, Shawn},
  booktitle={2020 IEEE Winter Conference on Applications of Computer Vision (WACV)}, 
  title={Cross-Time and Orientation-Invariant Overhead Image Geolocalization Using Deep Local Features}, 
  year={2020},
  volume={},
  number={},
  pages={2501-2509},
  doi={10.1109/WACV45572.2020.9093403}}

@ARTICLE{10422983,
  author={Liu, Huan and Li, Wei and Jia, Wen and Sun, Hong and Zhang, Mengmeng and Song, Lujie and Gui, Yuanyuan},
  journal={IEEE Transactions on Geoscience and Remote Sensing}, 
  title={Clusterformer for Pine Tree Disease Identification Based on UAV Remote Sensing Image Segmentation}, 
  year={2024},
  volume={62},
  number={},
  pages={1-15}
  }

@ARTICLE{10751785,
  author={Liu, Huan and Li, Wei and Xia, Xiang-Gen and Zhang, Mengmeng and Guo, Zhengqi and Song, Lujie},
  journal={IEEE Transactions on Image Processing}, 
  title={SegHSI: Semantic Segmentation of Hyperspectral Images With Limited Labeled Pixels}, 
  year={2024},
  volume={33},
  number={},
  pages={6469-6482}
}

@article{YI2023109019,
title = {UAVformer: A Composite Transformer Network for Urban Scene Segmentation of UAV Images},
journal = {Pattern Recognition},
volume = {133},
pages = {109019},
year = {2023},
issn = {0031-3203},
author = {Shi Yi and Xi Liu and Junjie Li and Ling Chen}
}

@article{YI2025661,
title = {An interactive fusion attention-guided network for ground surface hot spring fluids segmentation in dual-spectrum UAV images},
journal = {ISPRS Journal of Photogrammetry and Remote Sensing},
volume = {220},
pages = {661-691},
year = {2025},
issn = {0924-2716},
author = {Shi Yi and Mengting Chen and Xuesong Yuan and Si Guo and Jiashuai Wang}
}

@article{YI2023112612,
title = {CCTseg: A cascade composite transformer semantic segmentation network for UAV visual perception},
journal = {Measurement},
volume = {211},
pages = {112612},
year = {2023},
issn = {0263-2241},
author = {Shi Yi and Junjie Li and Gang Jiang and Xi Liu and Ling Chen}
}

@INPROCEEDINGS{7789580,
  author={Kampffmeyer, Michael and Salberg, Arnt-Børre and Jenssen, Robert},
  booktitle={2016 IEEE Conference on Computer Vision and Pattern Recognition Workshops (CVPRW)}, 
  title={Semantic Segmentation of Small Objects and Modeling of Uncertainty in Urban Remote Sensing Images Using Deep Convolutional Neural Networks}, 
  year={2016},
  volume={},
  number={},
  pages={680-688}}

@ARTICLE{FCN,
  author={Shelhamer, Evan and Long, Jonathan and Darrell, Trevor},
  journal={IEEE Transactions on Pattern Analysis and Machine Intelligence}, 
  title={Fully Convolutional Networks for Semantic Segmentation}, 
  year={2017},
  volume={39},
  number={4},
  pages={640-651}}

@Article{rs13234902,
AUTHOR = {Chen, Guanzhou and Tan, Xiaoliang and Guo, Beibei and Zhu, Kun and Liao, Puyun and Wang, Tong and Wang, Qing and Zhang, Xiaodong},
TITLE = {SDFCNv2: An Improved FCN Framework for Remote Sensing Images Semantic Segmentation},
JOURNAL = {Remote Sensing},
VOLUME = {13},
YEAR = {2021},
NUMBER = {23},
ARTICLE-NUMBER = {4902}
}

@ARTICLE{10608051,
  author={Nguyen, Gia-Vuong and Huynh-The, Thien},
  journal={IEEE Geoscience and Remote Sensing Letters}, 
  title={Enhancing Aerial Semantic Segmentation With Feature Aggregation Network for DeepLabV3+}, 
  year={2024},
  volume={21},
  number={},
  pages={1-5}}

@ARTICLE{deeplabv1,
  author={Chen, Liang-Chieh and Papandreou, George and Kokkinos, Iasonas and Murphy, Kevin and Yuille, Alan L.},
  journal={IEEE Transactions on Pattern Analysis and Machine Intelligence}, 
  title={DeepLab: Semantic Image Segmentation with Deep Convolutional Nets, Atrous Convolution, and Fully Connected CRFs}, 
  year={2018},
  volume={40},
  number={4},
  pages={834-848}}

@ARTICLE{9392319,
  author={Girisha, S. and Verma, Ujjwal and Manohara Pai, M. M. and Pai, Radhika M.},
  journal={IEEE Journal of Selected Topics in Applied Earth Observations and Remote Sensing}, 
  title={UVid-Net: Enhanced Semantic Segmentation of UAV Aerial Videos by Embedding Temporal Information}, 
  year={2021},
  volume={14},
  number={},
  pages={4115-4127}}

@article{wang2025multi,
  title={Multi-Sensor Data Fusion for Coastal Boundary Detection by Res-U-Net Implementation Using High-Resolution UAV Imagery},
  author={Wang, Qin and Kavhiza, Nyasha J and Islam, Fakhrul and Huqqani, Ilyas Ahmad and Abbas, Mohsin and Barman, Sanjoy},
  journal={IEEE Journal of Selected Topics in Applied Earth Observations and Remote Sensing},
  year={2025},
  publisher={IEEE}
}

@Article{isprs-annals-V-1-2022-15-2022,
AUTHOR = {El Amrani Abouelassad, S. and Rottensteiner, F.},
TITLE = {VEHICLE INSTANCE SEGMENTATION WITH ROTATED BOUNDING BOXES IN UAV IMAGES USING CNN},
JOURNAL = {ISPRS Annals of the Photogrammetry, Remote Sensing and Spatial Information Sciences},
VOLUME = {V-1-2022},
YEAR = {2022},
PAGES = {15--23}
}

@ARTICLE{9466361,
  author={Kang, Jian and Fernandez-Beltran, Ruben and Wang, Zhirui and Sun, Xian and Ni, Jingen and Plaza, Antonio},
  journal={IEEE Transactions on Geoscience and Remote Sensing}, 
  title={Rotation-Invariant Deep Embedding for Remote Sensing Images}, 
  year={2022},
  volume={60},
  number={},
  pages={1-13}}

@ARTICLE{10339376,
  author={Cao, Qinglong and Chen, Yuntian and Ma, Chao and Yang, Xiaokang},
  journal={IEEE Transactions on Geoscience and Remote Sensing}, 
  title={Few-Shot Rotation-Invariant Aerial Image Semantic Segmentation}, 
  year={2024},
  volume={62},
  number={},
  pages={1-13}}

@article{mo2024achieving,
  title={Achieving Rotation Invariance in Convolution Operations: Shifting from Data-Driven to Mechanism-Assured},
  author={Mo, Hanlin and Zhao, Guoying},
  journal={arXiv preprint arXiv:2404.11309},
  year={2024}
}

@article{mo2024ric,
  title={RIC-CNN: rotation-invariant coordinate convolutional neural network},
  author={Mo, Hanlin and Zhao, Guoying},
  journal={Pattern Recognition},
  volume={146},
  pages={109994},
  year={2024},
  publisher={Elsevier}
}

@article{mitton2021rotation,
  title={Rotation equivariant deforestation segmentation and driver classification},
  author={Mitton, Joshua and Murray-Smith, Roderick},
  journal={arXiv preprint arXiv:2110.13097},
  year={2021}
}

@inproceedings{e2cnn,
    title={{General E(2)-Equivariant Steerable CNNs}},
    author={Weiler, Maurice and Cesa, Gabriele},
    booktitle={Conference on Neural Information Processing Systems (NeurIPS)},
    year={2019},
}

@INPROCEEDINGS{11125636,
  author={Weijler, Lisa and Hermosilla, Pedro},
  booktitle={2025 International Conference on 3D Vision (3DV)}, 
  title={Efficient Continuous Group Convolutions for Local SE(3) Equivariance in 3D Point Clouds}, 
  year={2025},
  volume={},
  number={},
  pages={1372-1381}}

@article{yang2025group,
  title={Group Equivariant Convolutional Networks for Pathloss Estimation},
  author={Yang, Ziyue and Liu, Feng and Jin, Yifei and Vandikas, Konstantinos},
  journal={arXiv preprint arXiv:2511.17841},
  year={2025}
}

\setlength{\parskip}{0em}

\begin{IEEEbiography}
[{\includegraphics[width=1in,height=1.25in,clip,keepaspectratio]{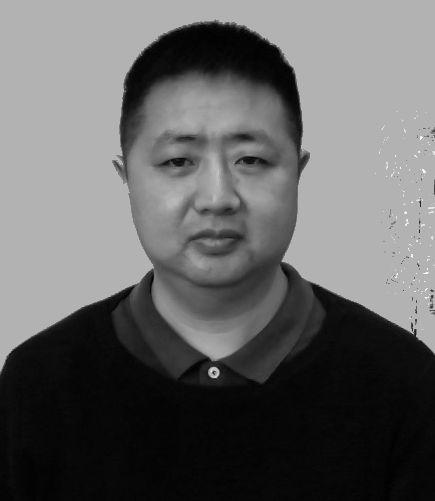}}]{Manduhu Manduhu}
 received the M.Sc. and Ph.D. degrees
in information engineering from Hiroshima University,
Japan, in 2010 and 2013, respectively.
He is currently a PostDoc researcher at the Drone Systems Lab of the University of the West of Scotland. His research interests include parallel algorithms, physics‑based simulation, and deep learning for object detection and segmentation.
\end{IEEEbiography}

\begin{IEEEbiography}
[{\includegraphics[width=1in,height=1.25in,clip,keepaspectratio]{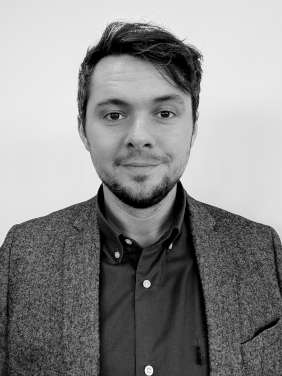}}]{Alexander Dow}
~received his B.S. in sound engineering and production from Birmingham City University,
Birmingham, UK in 2016 and his M.S. in digital signal and image processing from the 
University of Sussex, Brighton, UK in 2020. He is currently pursuing his Ph.D. in
drone-based computer vision at the University of the West of Scotland,
Lanarkshire, Scotland.
 
He currently works as a Post-Doctoral Research Assistant in the Drone Systems Lab at the University of the West of 
Scotland. His research interests include computer vision, artificial intelligence on embedded systems, LiDAR and hyperspectral imaging.
\end{IEEEbiography}


\begin{IEEEbiography}
[{\includegraphics[width=1in,height=1.25in,clip,keepaspectratio]{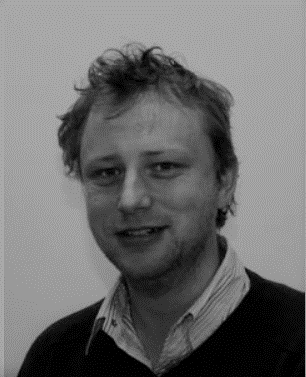}}]{Gerard Dooly}
received the B.Eng. degree from the Electronic and Computer Engineering Department, University of Limerick, in 2003, and the Ph.D. degree from the Optical Fibre Sensors Research Centre, 
University of Limerick, in 2008, on the topic “An Optical Fibre Sensor for the Measurement of Hazardous Emissions from Land Transport Vehicles.”
For over ten years, he has worked extensively in field robotics at UL. His research interests include real-time 3-D reconstruction, machine vision, machine learning, optical fibre sensors, subsea structural health monitoring, 
teleoperation, and automated docking and intervention. He is involved in robotics for harsh environments in offshore setting and is developing systems to address beyond visual line of sight operations for UAS. 
He is focused on the design and development of robotics and has engaged in numerous field operations and survey missions both in Ireland and on the continent. Some of his recent offshore operations involved environmental sensing, 
anti-mine countermeasure ops, remote UAS for incident response, archaeological survey, and hybrid long range UAS technologies.
\end{IEEEbiography}

\begin{IEEEbiography}
[{\includegraphics[width=1in,height=1.25in,clip,keepaspectratio]{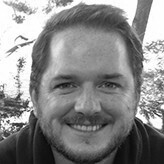}}]{James Riordan} 
~received the B.Eng. degree in electronic engineering specializing in aircraft simulation
systems and the Ph.D. degree in real-time processing of acoustic signals from the University of
Limerick, Ireland.
 
Currently, he is a Full Professor with the University of the West of Scotland, U.K.,
where he is also the Director of the Drone Systems Laboratory. He is also a Principal Investigator
of multiple research projects funded by European
Commission and U.K. Research and Innovation. His research interests include artificial intelligence,
computer vision, and sensing methods to extend the safe and sustainable
application of autonomous vehicles in land, air, and sea.
\end{IEEEbiography}

\end{document}